\pdfoutput=1

\documentclass[11pt]{article}

\usepackage[preprint]{acl}

\usepackage{times}
\usepackage{latexsym}

\usepackage[T1]{fontenc}

\usepackage[utf8]{inputenc}

\usepackage{microtype}

\usepackage{inconsolata}

\usepackage{graphicx}

\usepackage{booktabs}
\usepackage{siunitx}
\usepackage{xparse}
\usepackage{multirow}
\usepackage{listings}
\usepackage{float}
\usepackage{makecell} 
\usepackage{afterpage}
\usepackage{float}

\title{Relation Extraction or Pattern Matching? Unravelling the Generalisation Limits of Language Models for Biographical RE}

\author{
  \textbf{Varvara Arzt\textsuperscript{1,2}},
  \textbf{Allan Hanbury\textsuperscript{1}%
    \begingroup
    \hypersetup{linkcolor=black}%
    \thanks{Supervised by}%
    \endgroup},
  \textbf{Michael Wiegand\textsuperscript{3}},
  \textbf{Gábor Recski\textsuperscript{1}},
  \textbf{Terra Blevins\textsuperscript{4}\footnotemark[\value{footnote}]}
\\
  \textsuperscript{1}Faculty of Informatics, TU Wien,
  \textsuperscript{2}D!ARC, University of Klagenfurt,\\
  \textsuperscript{3}Digital Philology, University of Vienna,
\\
  \textsuperscript{4}Khoury College of Computer Sciences, Northeastern University
\\
    Correspondence: \texttt{varvara.arzt@tuwien.ac.at}
}
  
\begin{document}
\maketitle
\begin{abstract}
Analysing the generalisation capabilities of relation extraction (RE) models is crucial for assessing whether they learn robust relational patterns or rely on spurious correlations. Our cross-dataset experiments find that RE models struggle with unseen data, even within similar domains. Notably, higher intra-dataset performance does not indicate better transferability, instead often signaling overfitting to dataset-specific artefacts. Our results also show that data quality, rather than lexical similarity, is key to robust transfer, and the choice of optimal adaptation strategy depends on the quality of data available: while fine-tuning yields the best cross-dataset performance with high-quality data, few-shot in-context learning (ICL) is more effective with noisier data. However, even in these cases, zero-shot baselines occasionally outperform all cross-dataset results. Structural issues in RE benchmarks, such as single-relation per sample constraints and non-standardised negative class definitions, further hinder model transferability. We release our dataset splits with sample IDs and code for reproducibility.\footnote{\url{https://github.com/kleines-gespenst/re_cross_dataset}}
\end{abstract}

\section{Introduction}

Relation extraction (RE) is the core information extraction task of identifying the semantic relationship between entities in text. Traditional RE evaluations rely predominantly on in-distribution testing, but this approach often overestimates true model performance by implicitly assuming that individual datasets wholly represent the underlying task \citep{linzen-2020-accelerate,kovatchev-lease-2024-benchmark}. While model generalisation has gained increasing attention in NLP, RE remains relatively unexplored in this context (\S~\ref{sec:related_work}). 

However, understanding RE generalisation to out-of-distribution (OOD) data is crucial both for the task itself as well as for the robust application of RE systems in downstream tasks like question answering and knowledge-base population \citep{bassignana-plank-2022-crossre}. Given the popularity of representing internal language model (LM) knowledge as relational triples \citep{geva-etal-2023-dissecting,hernandez2024linearity}, building robust RE systems beyond the mere memorisation of dataset-specific patterns may also be key to more interpretable and trustworthy models.

This paper systematically analyses how well RE systems generalise across datasets focusing on sentence-level RE. 
Due to the limited relation overlap in popular RE datasets, we focus our experiments on biographical relations, which are pervasive in RE settings; this also allows us to include a domain-specific dataset for grounded analysis (\S~\ref{sec:data}). Through our cross-dataset experiments, this paper makes the following contributions: 

\begin{itemize}
\item We document key challenges in analysing RE generalisation, including inconsistent relation schemas and highly imbalanced class distributions (\S~\ref{sec:data}), as well as propose methods for overcoming these issues.

\item We find that strong in-distribution RE performance often masks fundamental generalisation failures, with models that excel on intra-dataset evaluations frequently failing to transfer effectively (\S~\ref{sec:results}). 

\item Our cross-dataset analysis suggests that how data is annotated influences the best adaptation method: in our experiments, fine-tuning achieves better cross-dataset performance for manually annotated data, while few-shot in-context learning (ICL) performs better on distantly supervised data. However, zero-shot prompting outperforms all cross-dataset methods in some settings (\S~\ref{sec:cross_dataset_res}). 

\item We identify structural issues in current RE benchmarks that lead to generalisation errors, including single-relation constraints, external knowledge reliance, and coverage biases (\S~\ref{sec:performance_analysis}). 
\end{itemize}

These findings reveal that while current RE systems achieve high in-distribution results, their cross-dataset performance shows critical gaps in genuine relation understanding, limiting their real-world applicability.

\section{Related Work}\label{sec:related_work}

\subsection{Approaches for RE}

RE is traditionally framed as a classification task, tackled via either a pipeline approach—where subtasks like named entity recognition (NER), coreference resolution, and relation classification (RC) are performed sequentially—or a joint model that processes them simultaneously \citep{taille-etal-2020-lets,bassignana-plank-2022-mean,saini-etal-2023-90}. It is further categorised into sentence- \citep{alt-etal-2020-tacred,10.1145/3477495.3531742} and document-level RE \citep{yao-etal-2019-docred,meng-etal-2024-robustness}. 

Since the introduction of BERT \citep{devlin-etal-2019-bert}, encoder-based models have dominated RE due to their bidirectional attention mechanism, which effectively captures context for classification tasks \citep{alt-etal-2020-tacred,10.1145/3477495.3531742}. However, the rise of autoregressive models has led to increasing adoption of decoder-based architectures to RE \citep{wang-etal-2022-deepstruct,sun-etal-2023-text,xu-etal-2023-unleash,liu-etal-2024-unleashing-power,efeoglu2024relationextractionfinetunedlarge}. While encoder-decoder models have been explored \citep{huguet-cabot-navigli-2021-rebel-relation,10164909}, our experiments focus on the dominant encoder-only and decoder-only architectures for RE.

\subsection{Generalisation Capabilities of RE Models}

Recent work advocates for transparent evaluation \citep{neubig-etal-2019-compare,explainaboard_custom} and OOD testing \citep{linzen-2020-accelerate,allenzhu2024physicslanguagemodels31,qi-etal-2023-preserving} to assess model robustness. Common strategies include cross-dataset \citep{antypas-camacho-collados-2023-robust,jang-frassinelli-2024-generalizable} and cross-domain \citep{fu-etal-2017-domain,Liu2020CrossNEREC,bassignana-plank-2022-crossre,calderon-etal-2024-measuring} experiments, as well as testing on perturbed and adversarial sets \citep{wu-etal-2019-errudite,gardner-etal-2020-evaluating,goel-etal-2021-robustness,rusert-etal-2022-robustness}. 

Recent studies have explored various ways to improve RE model robustness. \citet{bassignana-plank-2022-crossre} introduce a cross-domain RE dataset with broad relation types, while \citet{meng-etal-2024-robustness} and \citet{chen-etal-2023-models} evaluate state-of-the-art (SOTA) document-level RE models on perturbed test sets. \citet{chen-etal-2023-models} reveal that even when models predict correctly, they often rely on spurious correlations, showing vulnerability to minor evaluation shifts. To reduce dependence on mere pattern matching, \citet{allenzhu2024physicslanguagemodels31} propose augmenting training data with synthetic samples reformulated by an auxiliary model. Most closely related to our work, \citet{bassignana-plank-2022-mean} analysed cross-dataset model transfer for scientific RE with large data overlap, while our work examines cross-dataset generalisation across models for general-purpose RE and analyses factors influencing transfer considering datasets with no overlap. 

\section{Methodology}\label{sec:data}
We assess the  RE systems' robustness by evaluating their OOD performance. Standard in-distribution evaluations may overestimate RE performance \citep{linzen-2020-accelerate}, as models can exploit spurious cues rather than learning genuine RE task \citep{chen-etal-2023-models,meng-etal-2024-robustness,arzt-hanbury-2024-beyond}.

To systematically evaluate generalisation capabilities of RE models, we conduct both intra- and cross-dataset experiments. The intra-dataset experiments act as a control, evaluating RE models on data drawn from the distribution used for model adaptation, while the cross-dataset experiments measure model robustness with OOD test sets derived from a different RE dataset. 

For our experiments, we use three sentence-level RE datasets: TACRED-RE \citep{alt-etal-2020-tacred}, NYT \citep{10.1007/978-3-642-15939-8_10}, and Biographical \citep{10.1145/3477495.3531742}. While TACRED-RE and NYT are general-purpose RE datasets, we focus only on biographical relations, or relations describing aspects of an individual's life like \textit{place\_of\_birth} or \textit{children} for two key factors: (1) TACRED-RE and NYT share six biographical relations but only two non-biographical ones, and (2) focusing on biographical relations allows for additional cross-dataset evaluations with the Biographical dataset, which only contains biographical relations. This setup thus allows us to compare the generalisation of two popular RE datasets in a third, held-out evaluation setting.

\subsection{Data}\label{sec:datasets}
We now briefly describe three RE datasets used.

\begin{figure}[ht]
    \centering
    \includegraphics[width=\linewidth]{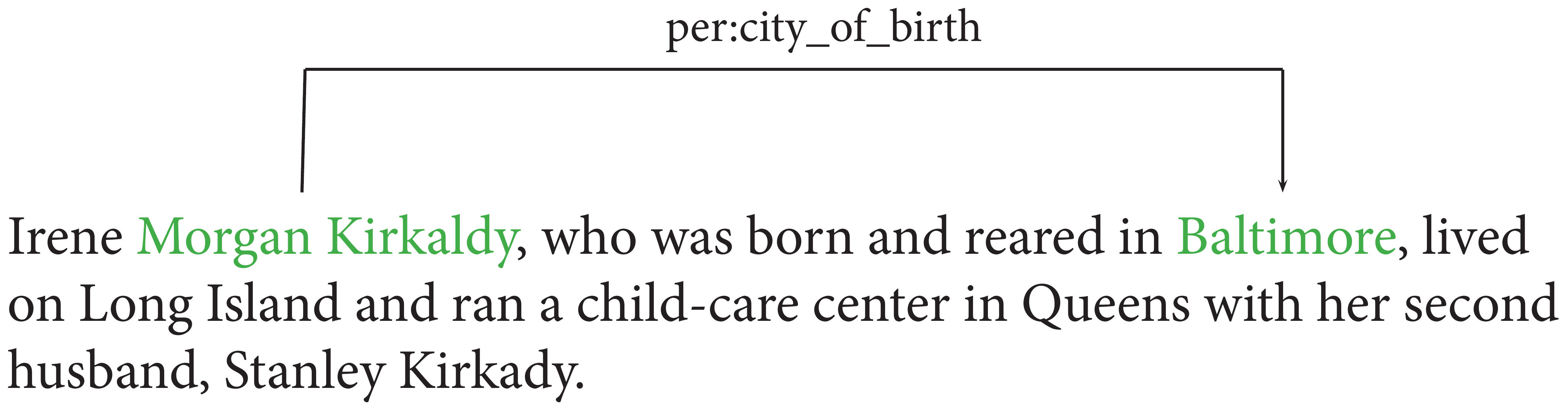}
    \caption{TACRED-RE example \citep{zhang-etal-2017-position}.}
    \label{fig:tacred_example}
\end{figure}

\textbf{TACRED-RE} \citep{alt-etal-2020-tacred} is a general-purpose RE dataset with 41 relations and a `no\_relation' class.\footnote{Licensed by the Linguistic Data Consortium (LDC).} It contains over 106k instances but is highly imbalanced, with $\sim$80\% labeled as `no\_relation'. Built from English newswire and web text, it is a revised version of TACRED \citep{zhang-etal-2017-position}, with challenging samples re-annotated by professional annotators to reduce noise from crowdsourcing. Experimental results show improved performance on TACRED-RE compared to TACRED (Appendix Tables~\ref{tacred-results},~\ref{tacred-re-results}), leading to its use in our experiments. Figure~\ref{fig:tacred_example} shows a TACRED-RE example. We focus on its 26 biographical relations, including `no\_relation' (Appendix Table~\ref{tacred-re-table}). 


%

\begin{figure}[ht]
    \centering
    \includegraphics[width=\linewidth]{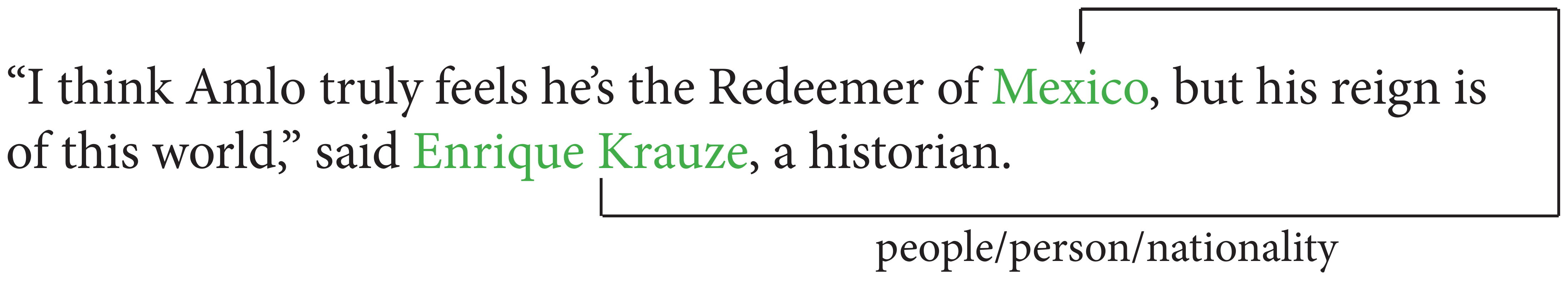}
    \caption{NYT example.}
    \label{fig:nyt_example_1rel}
\end{figure}

 \textbf{NYT} \citep{10.1007/978-3-642-15939-8_10} is a general-purpose RE dataset with 24 relations and a `None' class. It contains over 266k sentences, with 64\% labeled as `None' and half of positive instances containing a single dominant relation, `/location/location/contains'.\footnote{Available at \url{https://github.com/INK-USC/ReQuest}.} NYT was constructed via distant supervision, by applying \texttt{Freebase} \citep{10.1145/1376616.1376746} as external supervision to New York Times articles \citep{Sandhaus2008Nyt}. Figure~\ref{fig:nyt_example_1rel} shows an NYT example. We focus on its subset with 7 biographical relations, including a `None' class (Table~\ref{nyt-table}, Appendix).

\begin{figure}[ht]
    \centering
    \includegraphics[width=\linewidth]{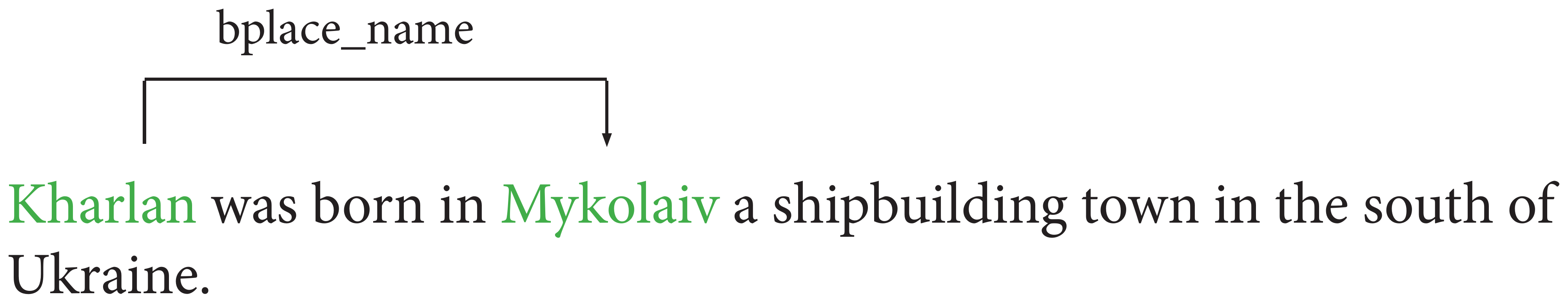}
    \caption{Biographical example.}
    \label{fig:bio_example}
\end{figure}

\textbf{Biographical} is an RE dataset for the biographical text domain, with 10 relation types \citep{10.1145/3477495.3531742}. Built from Wikipedia articles on prominent individuals and containing 346,257 instances, Biographical was created using a semi-supervised approach.\footnote{We use the \textit{m2\_normal\_final1} version.} Named entities were automatically extracted using \texttt{spaCy} \citep{Honnibal_spaCy_Industrial-strength_Natural_2020} and \texttt{Stanford CoreNLP} \citep{manning-etal-2014-stanford}. \texttt{Wikipedia} sentences with these entities were matched with \texttt{Pantheon} and \texttt{Wikidata} to automatically infer relations. Figure~\ref{fig:bio_example} shows an example\footnote{Sample ID `mS1/18860978' shows punctuation revomal by \citet{10.1145/3477495.3531742} that may affect model comprehension.} from Biographical. Statistics for Biographical, downsampled to match the size of the TACRED-RE and NYT subsets, appear in Appendix Table~\ref{nyt-table}. 

\subsection{Cross-Dataset Comparison: Challenges}\label{sec:challenges}
\paragraph{Single Relation per Sample:} Both TACRED-RE and Biographical restrict each sample to at most two entities and one relation, even when multiple relations exist in a sentence. For instance, the TACRED-RE example in Figure~\ref{fig:tacred_example} is labeled with `per:city\_of\_birth' but also contains `per:stateorprovinces\_of\_residence', `per:employee\_of', and `per:spouse', all within TACRED-RE's relation set. This constraint may confuse a model trained on such data, as it enforces a single-label assignment. While NYT better reflects real-world scenarios by allowing multiple relations per sentence, we filtered it to two-entity, single-relation samples for fair cross-dataset comparison, retaining only examples like in Figure~\ref{fig:nyt_example_1rel}.



\paragraph{Unclear `Negative' Class:} 

Clear negative samples—instances with entities but no meaningful relation—are crucial for RE. While TACRED-RE has an explicit `no\_relation' class, NYT's `None' class lacks clear definition \citep{10.1007/978-3-642-15939-8_10}, potentially confusing models about whether it indicates absence of predefined relations or any relation. Similarly, Biographical lacks an explicit negative class, using instead an `Other' class for unspecified relations \citep{10.1145/3477495.3531742}, exemplified in Appendix Figure~\ref{fig:bio_example_other}. This inconsistency between choosing a `no\_relation' versus a `none\_of\_the\_above' class in RE benchmarks highlights the general challenge of consistently defining the boundaries between presence and absence of semantic relations in text \citep{bassignana-plank-2022-mean}. 

\paragraph{Expected Factual Knowledge:} The design of RE datasets influences whether models genuinely learn RE or rely on dataset-specific cues. NYT's distant supervision approach incorporates Freebase-derived relations not stated in text requiring external world knowledge rather than textual evidence, as shown in Figure~\ref{fig:nyt_example_1rel} where the text lacks explicit information about Enrique Krause’s nationality---such annotations extend beyond RE's scope and corrupt models trained on such data. Similarly, although manually curated, TACRED-RE encompasses relations like `per:city\_of\_birth', which require factual knowledge from a model, limiting generalisation to instances seen during adaptation.

\subsection{Cross-Dataset Label Overlap}
Following \citet{bassignana-plank-2022-mean}, we manually analysed instances for each relation in TACRED-RE, NYT, and Biographical to establish a cross-dataset label mapping. Appendix Table~\ref{tab:overlap_nyt_tacred} shows six overlapping biographical relations between NYT and TACRED-RE, with twelve fine-grained TACRED-RE relations mapping to six broader NYT labels (e.g., NYT's `place\_of\_birth' encompasses three TACRED-RE birth location relations). Treating Biographical's `Other' class as negative—supported by manual analysis of 30 random instances showing negative rather than unspecified relations—we find four overlapping relations across three datasets (Appendix Table~\ref{tab:overlap_biographical_tacred_nyt}). NYT and Biographical share these same four relations (Appendix Table~\ref{tab:nyt_bio_mapping}), while TACRED-RE and Biographical share nine relations (Appendix Table~\ref{tab:bio_tacred_overlap}).

\section{Experiments}\label{sec:experiments}
\subsection{Data Format and Standardisation}\label{sec:data_format}

To enable cross-dataset evaluations and focus exclusively on \textit{relation classification}, we standardise our data using unified format with entity spans marked as \texttt{<e1>}head entity\texttt{</e1>} and \texttt{<e2>}tail entity\texttt{</e2>}.

To address class imbalance, we randomly downsampled negative instances across three datasets to balance the number of positive and negative instances, and downsampled Biographical ($\sim$350k) to match other biographical subset sizes for fair comparison (Appendix~\ref{sec:dataset_statistics}). For cross-dataset evaluation, we mapped TACRED-RE's fine-grained relations to broader NYT labels (Appendix~\ref{sec:data_processing}).





\subsection{Model Selection, Training, and Evaluation}
We consider two types of models: an encoder-only (DeBERTa-v3-large 304M; \citet{he2021deberta}) and a decoder-only model (an instruction-tuned LLaMA 3.1 8B; \citet{grattafiori2024llama3herdmodels}). For DeBERTa, we follow common practice for RE encoder models: we mark entity spans with custom markers and use the concatenated hidden states of the entity start tokens, \texttt{<e1>} and \texttt{<e2>}, as inputs for the classification head \citep{baldini-soares-etal-2019-matching}. We also evaluate three SOTA systems (or replications of these systems) on our biographical test sets, one for each considered dataset, in order to compare our models to prior work. We provide more details on our replication study in Appendix \ref{sec:replication_sota}.

We employ two commonly used model adaptation strategies for RE: fine-tuning and in-context learning (ICL). Specifically, we consider direct fine-tuning with DeBERTa, fine-tuning LLaMA using low-rank adaptation \citep[LoRA;][]{hu2022lora}, and zero-shot and five-shot ICL \citep{10.5555/3495724.3495883} with LLaMA. For few-shot ICL, we perform five runs with different demonstration sets to account for demonstration sensitivity \citep{zhang-etal-2022-active,webson-pavlick-2022-prompt,lu-etal-2022-fantastically}. However, due to computational constraints, fine-tuning experiments are limited to a single run. For NYT and TACRED-RE, we conduct experiments in two adaptation settings: adaptation on all biographical relations in each dataset (Appendix Tables~\ref{tacred-re-table} and~\ref{nyt-table}) and adaptation on only overlapping relations (Appendix Table~\ref{tab:overlap_nyt_tacred}). This applies to both fine-tuning and ICL, where zero-shot prompts and few-shot demonstrations are selected accordingly.

We then perform two types of evaluations: \textbf{intra-dataset}, where models are evaluated on the same dataset they were adapted to; and \textbf{cross-dataset}, where the adapted models were tested on OOD data to assess their generalisation. Further implementation details, including hyperparameter settings, and prompting details, are provided in Appendix~\ref{sec:model_implementation}.

\section{Results}\label{sec:results}

\begin{table*}[t]
\centering
\small
\setlength{\tabcolsep}{5pt}
\begin{tabular}{l|l|l|c|c|c|c}
\hline
\textbf{Model} & \textbf{Setting} & \textbf{Dataset} & \multicolumn{2}{c|}{\textbf{Intra-Dataset}} & \multicolumn{2}{c}{\textbf{Cross-Dataset}} \\
\hline
 &  & & \textbf{Shared Labels} & \textbf{Dataset Labels} & \textbf{NYT} & \textbf{TACRED-RE} \\
\hline
\makecell{DeBERTa-v3\\large 304M} & Fine-tuned on & NYT & 0.81 & 0.81 & -- & 0.26 \\
 &  & TACRED-RE & 0.72 & 0.62 & 0.53 & -- \\
\hline
LLaMA 3.1 8B & Fine-tuned on & NYT & \textbf{0.87} & 0.87 & -- & 0.45 \\
 &  & TACRED-RE & \textbf{0.82} & 0.76 & \textbf{0.62} & -- \\
\hline
\hline
LLaMA 3.1 8B & Zero-Shot & NYT & 0.31 & 0.31 & -- & -- \\
 &  & TACRED-RE & 0.58 & 0.37 & -- & -- \\
\hline
LLaMA 3.1 8B & 5-Shot & NYT & 0.45 ± 0.07 & 0.45 ± 0.07 & -- & \textbf{0.52 ± 0.06} \\
 &  & TACRED-RE & 0.63 ± 0.06 & 0.43 ± 0.07 & 0.39 ± 0.02 & -- \\
\hline
\end{tabular}
\caption{Macro F1-scores for intra- and cross-dataset predictions on six overlapping relations. Results include shared and dataset-specific labels, with models adapted on all biographical relations via fine-tuning or ICL. Best intra- and cross-dataset results are in bold.}
\label{tab:results_tacrev_nyt}
\end{table*}

\begin{table*}[t]
\centering
\small
\setlength{\tabcolsep}{2pt}
\begin{tabular}{l|l|l|c|c|c}
\hline
\textbf{Model} & \textbf{Setting} & \textbf{Dataset} & \textbf{Full Overlap} & \textbf{Overlap w. NYT} & \textbf{Overlap w. TACRED-RE} \\
\hline
DeBERTa-v3-large 304M & Fine-tuned on & NYT & 0.48 & 0.48 & -- \\
 &  & TACRED-RE & 0.62 & -- & 0.70 \\
  &  & Biographical & \textbf{0.80} & 0.80 & 0.81 \\
\hline
LLaMA 3.1 8B & Fine-tuned on & NYT & 0.30 & 0.30 & -- \\
 &  & TACRED-RE & \textbf{0.69} & -- & 0.70 \\
 &  & Biographical & 0.79 & 0.79 & 0.74 \\
\hline
\hline
LLaMA 3.1 8B & Zero-Shot & Biographical & 0.24 & 0.24 & 0.35 \\
\hline
LLaMA 3.1 8B & 5-Shot & NYT & 0.48 ± 0.04 & 0.48 ± 0.04 & -- \\
 &  & TACRED-RE & 0.51 ± 0.04 & -- & 0.58 ± 0.02 \\
 &  & Biographical & 0.53 ± 0.05 & 0.53 ± 0.05 & 0.54 ± 0.03 \\
\hline
\end{tabular}
\caption{Evaluation on Biographical Dataset (macro F1-scores). Models adapted on all biographical relations through fine-tuning or ICL. Best intra- and cross-dataset results on full overlap are in bold.}
\label{tab:results_bio_overlap_macro}
\end{table*}

\paragraph{Overview of Reported Results:} Table~\ref{tab:results_tacrev_nyt} shows intra- and cross-dataset results for NYT and TACRED-RE on six overlapping relations, using models adapted on all biographical relations. For TACRED-RE, which maps its 12 fine-grained labels to NYT's shared label space, we report both dataset-specific and shared label results, denoted as `Dataset Labels' and `Shared Labels' in Table~\ref{tab:results_tacrev_nyt}. Table~\ref{tab:results_bio_overlap_macro} shows model generalisation to Biographical across three overlapping relation sets:  (1) four relations shared by all datasets, (2) same four relations shared between NYT and Biographical, and (3) nine TACRED-RE/Biographical shared relations. Models were adapted on each dataset's full biographical relations, with Biographical's intra-dataset results for comparison. We focus on results with models adapted on the full overlap, as they perform similarly to those adapted only on overlap (Appendix Table~\ref{tab:results_tacrev_nyt_6rel}) while better reflecting real-world settings. Cross-dataset experiments with Biographical as training source appear in Appendix~\ref{sec:cross_dataset_bio}. Given Biographical's ambiguous `Other' class (\S~\ref{sec:challenges}), we use it only for evaluation in the main paper.

While we primarily focus on macro F1 to address class imbalance, we report micro F1 (commonly reported for RE tasks) for our fine-tuned models in Table \ref{tab:sota}; additional experimental results, including per-class breakdowns, appear in Appendix~\ref{sec:appendix_results}. 

\subsection{Intra-Dataset Results}\label{sec:intra-dataset_res}

We evaluate our RE models on their training data distribution for comparison of cross-dataset generalisation. Unsurprisingly, we find that fine-tuning performs best for intra-dataset evaluations: fine-tuned LLaMA outperforms DeBERTa on TACRED-RE and NYT (Table~\ref{tab:results_tacrev_nyt}), while DeBERTa outperforms LLaMA on Biographical (Table~\ref{tab:results_bio_overlap_macro}). 

Our ICL experiments similarly show expected results, with the five-shot prompting moderately outperforming zero-shot prompting but underperforming full model fine-tuning. For Biographical, this few-shot ICL gain over zero-shot is significant, increasing from $0.24$ to $0.53\pm 0.05$ with five demonstrations (Table \ref{tab:results_bio_overlap_macro}).

We observe different performance trends within TACRED-RE and NYT intra-dataset evaluations. While fine-tuning yields higher intra-dataset performance on NYT than on TACRED-RE, this trend flips for the zero- and few-shot ICL settings, with prompting on NYT performing significantly \textit{worse} despite TACRED-RE's finer-grained relation schema. This likely stems from differing data quality between datasets: the noisy labeling during NYT creation \citep{yaghoobzadeh-etal-2017-noise} likely leads to over-fitting during fine-tuning \citep{tanzer-etal-2022-memorisation} (rather than learning robust relational patterns), but harms model generalisation to NYT when not fine-tuned for that data distribution.

\begin{table}[t]
\centering
\tiny
\begin{tabular}{l|c|c|c|c|c|c}
\hline
\multirow{2}{*}{\textbf{Dataset}} & \multicolumn{2}{c|}{\textbf{SOTA}$^*$} & \multicolumn{2}{c|}{\makecell{\textbf{DeBERTa-v3} \\ \textbf{large 304M}}} & \multicolumn{2}{c}{\makecell{\textbf{LLaMA 3.1} \\ \textbf{8B}}} \\
\cline{2-7}
 & \textbf{Macro} & \textbf{Micro} & \textbf{Macro} & \textbf{Micro} & \textbf{Macro} & \textbf{Micro} \\
\hline
NYT & 0.87 & \underline{0.90} & 0.84 & 0.93 & 0.89 & 0.94 \\
TACRED-RE & 0.78 & \underline{0.87} & 0.71 & 0.85 & 0.78 & 0.89 \\
Biographical & \underline{0.87} & 0.93 & 0.83 & 0.92 & 0.75 & 0.91 \\
\hline
\end{tabular}
\caption{F1 score comparison of our models (trained on biographical subsets) vs. SOTA systems (trained on full datasets) when evaluated on biographical test sets. \underline{Underlined} values indicate which metric (macro or micro F1) was originally reported in SOTA$^*$ papers from \citet{orlando-etal-2024-relik} (NYT), \citet{zhou-chen-2022-improved} (TACRED-RE), and \citet{10.1145/3477495.3531742} (Biographical).}
\label{tab:sota}
\end{table}

\paragraph{Comparison with SOTA:} As our cross-dataset analysis focuses on biographical relations, direct comparisons with prior work are challenging as they rarely report per-class results. Thus, we perform a replication study to re-evaluate prior work on our biographical evaluation sets, comparing SOTA models—trained using full training and validation datasets—against our models fine-tuned exclusively on biographical subsets.

Table~\ref{tab:sota} shows that our models perform well on all three biographical test sets, achieving competitive or superior results to existing RE systems despite their more limited training data.

In most settings, our models outperform prior work, with one exception: on the Biographical dataset, our DeBERTa model trails the model with entity tags introduced in \citet{10.1145/3477495.3531742} by 4 and 1 points in macro/micro F1. This gap is likely attributable to training data disparity (our 20K vs. their 346K), particularly since both approaches use identical text formatting with entity markers and entity token representations for the classification head. Replication details appear in Appendix~\ref{sec:replication_sota}.

\subsection{Cross-Dataset Results}\label{sec:cross_dataset_res}

We now turn to examining the cross-dataset generalisation of our RE systems. Unsurprisingly, we find that performance almost always declines with cross-dataset evaluations. However, models adapted on TACRED-RE exhibit relatively strong generalisation capabilities---the few exceptions of better cross-dataset performance stemming from TACRED-RE models applied to the Biographical dataset--while those adapted on NYT struggle to transfer effectively, likely due to dataset noise.


\paragraph{RE Models Struggle to Generalise across Datasets}

Cross-dataset evaluations (almost) always perform worse than the comparable intra-dataset experiment: NYT and TACRED-RE show substantial drops of 20-30 points, while Biographical exhibits a smaller decrease of $\sim$10 points for both full and TACRED-RE/Biographical relation overlap (Tables~\ref{tab:results_tacrev_nyt},~\ref{tab:results_bio_overlap_macro}). We also observe somewhat different performance trends across model and adaptation approaches from the intra-dataset experiments; while fine-tuning LLaMA on TACRED-RE achieves the best cross-dataset performance on NYT, the best TACRED-RE cross-dataset results are obtained using few-shot ICL with NYT demonstrations (rather than fine-tuning). However, these remain below the zero-shot TACRED-RE baseline.

The cross-dataset experiments on Biographical similarly perform worse than the corresponding intra-dataset experiments in most settings (Table~\ref{tab:results_bio_overlap_macro}); one notable exception is LLaMA prompted with five TACRED-RE examples, which outperforms the intra-dataset few-shot experiments on the TACRED-RE/Biographical label overlap. The best Biographical cross-dataset results are achieved with fine-tuning LLaMA on TACRED-RE, though this still underperforms intra-dataset fine-tuning.


\paragraph{NYT Models Generalise Worse than TACRED-RE Models}

NYT-adapted models exhibit significantly poorer generalisation than those adapted on TACRED-RE (Tables~\ref{tab:results_tacrev_nyt} and~\ref{tab:results_bio_overlap_macro}). For example, fine-tuning LLaMA with TACRED-RE surpasses zero- and cross-dataset ICL on NYT (by $\sim$30 and $\sim$20 points, respectively), while all cross-dataset experiments transferring from NYT to TACRED-RE underperform zero-shot evaluations with no cross-dataset signal. This is also clear from the evaluations on the held-out Biographical dataset, where transferring from TACRED-RE always performs better than NYT (and occasionally outperforms the intra-dataset performance). 


This performance gap is unlikely due to domain differences, as both datasets contain newspaper articles (TACRED-RE includes some NYT newspaper content without instance overlap), while Biographical uses Wikipedia. We attribute it to NYT's distant supervision annotations, which introduce noise and limit model robustness.
This is likely why LLaMA fine-tuned on NYT and evaluated on Biographical ($0.30$) underperforms DeBERTa ($0.48$; Table~\ref{tab:results_bio_overlap_macro})---the over-parametrised LLaMA exhibits stronger overfitting to NYT noise and generalises poorly to unseen data, a phenomenon also noted by \citet{pmlr-v162-liu22w} with corrupted training data.

\paragraph{Effect of Adaptation Strategy on Generalisation}\label{adaptation_strategy_effect}

While prior work suggests that ICL often generalises more effectively to OOD data than fine-tuning \citep{awadalla-etal-2022-exploring,10.1145/3582688,si2023prompting}, our results indicate this advantage depends heavily on data quality. With high-quality data like TACRED-RE, fine-tuning consistently achieves the best cross-dataset performance, surpassing few-shot ICL on both NYT and Biographical evaluations. In fact, TACRED-RE adaptations can even perform comparably to intra-dataset ones: LLaMA ($0.70$) fine-tuned on TACRED-RE achieves similar results to its Biographical intra-dataset performance ($0.74$) on the TACRED-RE/Biographical label overlap (Table~\ref{tab:results_bio_overlap_macro}). 

However, when adaptation data are noisy, as with NYT, few-shot ICL becomes a more effective strategy: few-shot ICL via NYT consistently performs better than fine-tuning on NYT for both TACRED-RE and Biographical evaluations (Tables~\ref{tab:results_tacrev_nyt},~\ref{tab:results_bio_overlap_macro}). This is likely because ICL limits the signal from noisy training data, which in turn reduces the overfitting to dataset-specific artefacts and catastrophic forgetting compared to fine-tuning \citep{tran-etal-2024-preserving,kotha2024understanding,10889361}.  

\section{Analysing RE Generalisation Failure Cases}\label{sec:performance_analysis}

\begin{figure}[t]
    \centering
    \includegraphics[width=\linewidth]
    {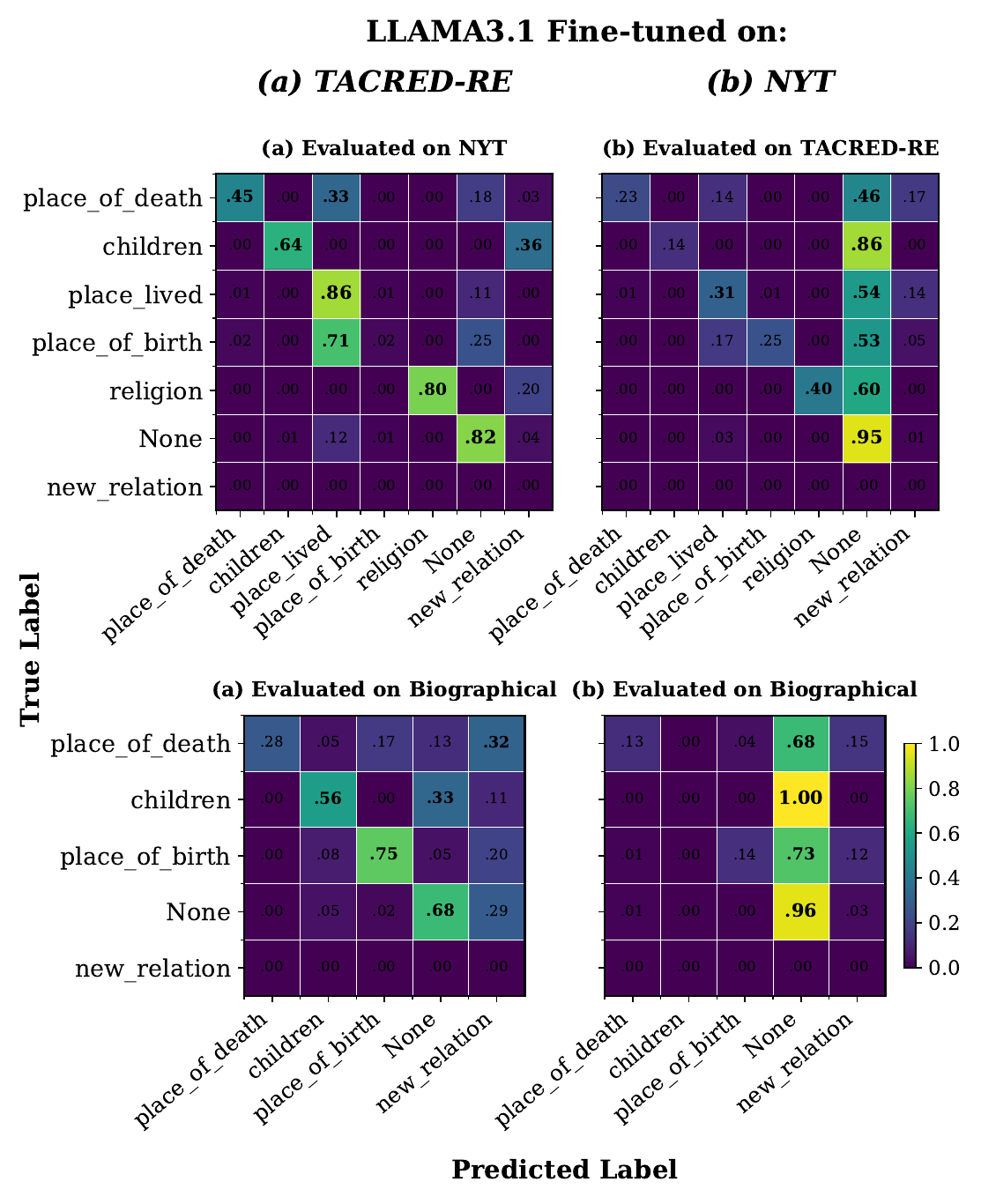}
    \caption{Confusion matrices comparing cross-dataset results for LLaMA fine-tuned on TACRED-RE/NYT.}
    \label{fig:conf_matrices}
\end{figure}



Given RE benchmarks' numerous relations and class imbalance, we analyse the strongest cross-dataset performing model, fine-tuned LLaMA, beyond aggregated metrics. We examine per-relation performance (Figure~\ref{fig:conf_matrices}) and qualitatively analyse 30 random misclassifications from four evaluation settings to identify their likely underlying causes.\footnote{More misclassification patterns in Appendix Table~\ref{tab:consolidated_misclassification_analysis}.} Through these analyses, we find the following causes of RE generalisation mistakes:

\paragraph{Effect of Noisy Supervision on Generalisation}

Figure~\ref{fig:conf_matrices} reveals that NYT-adapted models systematically overpredict the `None' class on both TACRED-RE and Biographical, with manual analysis showing misclassifications stem primarily from NYT's \textit{distant supervision} (Appendix Table~\ref{tab:consolidated_misclassification_analysis}) rather than vocabulary differences (Appendix Figure~\ref{fig:vocab_overlap_nyt_tacrev}) or domain shift. This issue is particularly evident in NYT’s location-based relations, where reliance on external knowledge leads to conflicting annotations that hinder pattern learning---for instance, ``\texttt{Henryk Tomaszewski} [...] died on Sunday at his home in \texttt{Warsaw}''\footnote{NYT instance ID: `/m/vinci8/data1/riedel/projects/relation\allowbreak/kb\allowbreak/nyt1/docstore\allowbreak/nyt-2005-2006.backup\allowbreak/1701917.xml.pb'} is labeled as birthplace despite clear evidence of death location. Similar issues arise with Biographical's \textit{semi-supervised} data, where models adapted on cleaner datasets like TACRED-RE fail to replicate ground truth labels that lack textual evidence. Notably, despite higher lexical overlap between Biographical and NYT (Appendix Figure~\ref{fig:vocab_overlap_bio}), TACRED-RE-adapted models perform better on Biographical, indicating that adaptation data quality matters more than lexical similarity.

\paragraph{Single Relation Constraint \& Negative Class}
The issue extends beyond noisy supervision to fundamental RE task design constraints. Models often detect valid but unlabeled relations (marked as `new\_relation' in Figure~\ref{fig:conf_matrices}), revealing limitations of single-relation per sample. This manifests in cases like ``\texttt{Gross}, who is himself Jewish [...] was sent to \texttt{Cuba}''\footnote{\label{foot:TACRED}TACRED-RE instance ID: `098f6f318be29eddb182'}, where only `place\_lived' is labeled while `religion' is omitted due to TACRED-RE's constraint. Also, unclear negative class affects cross-dataset evaluation: while Biographical's `Other' class is intended to cover undefined relations, our analysis reveals it contains both instances without meaningful relations and those with valid but undefined relations. This explains the high frequency of `new\_relation' predictions for Biographical when using LLaMA fine-tuned on TACRED-RE with their finer-grained relation schema (Figure~\ref{fig:conf_matrices}) and highlights the fundamental difficulty in defining boundaries between relation presence or absence. Despite this ambiguity, the class achieves strong cross-dataset performance when mapped to `None' ($0.78$ and $0.72$ for LLaMA fine-tuned on TACRED-RE and NYT (Appendix Tables~\ref{tab:bio-per-class-finetuned-tacrev} and~\ref{tab:bio-per-class-finetuned-nyt})), further underscoring the importance of distinguishing between `no\_relation' and `none\_of\_the\_above' cases \citep{bassignana-plank-2022-mean}.

\paragraph{Reliance on External Knowledge even in Manually Curated Datasets}
Even with high-quality manual annotations, RE often requires external knowledge and complex reasoning capabilities. Our analysis reveals this challenge manifests in two key ways: through implicit relations requiring inference, and through necessary world knowledge for entity interpretation. For example, in ``\texttt{Gross} [...] was sent to \texttt{Cuba} as a spy''\footref{foot:TACRED}, the NYT-adapted model predicts `None' instead of `place\_lived', failing to infer  that being sent somewhere as a spy implies residence. While detecting implicit relations is crucial \citep{geva-etal-2021-aristotle}, ensuring consistent and objective interpretation remains challenging.

Beyond implicit relations, models must also rely on world knowledge for basic entity understanding-as in cases like `Idaho businesswoman'\footnote{TACRED-RE instance ID: `098f6bd9fa786293e49d'}, where identifying entity types requires knowing Idaho as a location. TACRED-RE fine-grained relation schema further demonstrates this issue, where  even with world knowledge, distinguishing between relations like city of birth and state/province of birth can be ambiguous (e.g., whether New York refers to the city or state). As \citet{chen-etal-2023-models} note, even human annotators tend to rely on such prior knowledge despite the lack of rationales, motivating the need for finer-grained word evidence annotation.

\paragraph{Dataset Composition and Coverage Biases}
Analysis of the most frequent words across NYT and TACRED-RE shows a strong US-centric coverage bias, likely limiting generalisation to non-US contexts (Appendix Table~\ref{tab:top_words_nyt_tacrev_extended}). NYT also exhibits topical skews in specific relations, like religion being primarily associated with Islam, potentially biasing model relation representations.

Analysing part-of-speech distributions also reveals distinct patterns across all three datasets (Appendix Table~\ref{tab:pos_distribution}). While proper nouns dominate head and tail entities in all datasets (nearly 100\% in NYT), TACRED-RE shows more linguistic diversity with 17\% of head entities as pronouns and 17\% of tail entities as common nouns. Biographical, sourced from Wikipedia, contains a high proportion (26\%) of numerical tail entities, primarily dates. These compositional differences, along with TACRED-RE's longer, compound sentences and higher average entity distance ($\sim$12 tokens vs NYT's $\sim$8 tokens), most likely impact cross-dataset performance; NYT-adapted models struggle with these more complex patterns, which are absent from their training data (Appendix Table~\ref{tab:consolidated_misclassification_analysis}).

\section{Conclusion}

This work examines cross-dataset generalisation in language model-based RE systems in biographical settings. We find RE models struggle to generalise even within similar domains, with high intra-dataset performance potentially masking spurious overfitting rather than indicating genuine learning of relational patterns. Our empirical results suggest that data annotation quality can significantly influence transfer; specifically, we observe that datasets with human annotations provide notably better cross-dataset performance than those created through distant supervision. The best adaptation strategy also appears dataset-dependent, with fine-tuning yielding the best cross-dataset performance with TACRED-RE as the adaptation source while few-shot ICL appears to offer advantages with noisier data from our experiments. However, in some cases, a zero-shot baseline surpasses all cross-dataset results, further underscoring the limitations of current RE systems.


Our analysis also reveals several structural issues in all current RE benchmarks: (1) single-relation constraints that ignore other valid relations between entities in text, (2) the lack of a well-defined negative class with challenging samples (e.g., sentences containing commonly used tokens for relations like `born' or `died') to enforce deeper semantic understanding beyond pattern matching, and (3) limited diversity in data sources. 
These issues, compounded by inconsistent relation definitions and limited overlap across datasets, hinder meaningful evaluations of RE generalisation.

These findings thus highlight the need for more transparent evaluation beyond in-distribution testing and aggregated metrics, as limiting evaluation to these may not reflect genuine improvements in capturing relational patterns or account for class imbalance and the large number of relations in RE benchmarks. We see many promising directions for future work, including testing RE robustness on perturbed evaluation sets and applying interpretability methods to better understand how models infer relational knowledge.

\section*{Limitations}

Our cross-dataset analysis is limited to a particular set of biographical relations but reflects a broader challenge in RE evaluation where datasets, even covering the same domain, typically share a small relation overlap. We also constrain our analysis to single-relation examples: while, real-world scenarios often involve multiple relations per instance (and NYT allows multiple relations), we focused on single-relation setting for fair cross-dataset comparison, as TACRED-RE and Biographical are annotated with single relations. Similarly, we exclusively evaluate relation classification (RC) due to dataset constraints: TACRED-RE and Biographical assume a single relation triple per sentence, unlike real-world text where multiple relations can coexist. By focusing on RC with entity tags as guidance, we aim to minimise the prediction of other potential relations present in a sentence, but not between the specified entities.

The adaptation sets we used contain a large class imbalance due to the underlying distributions of the datasets, even after we perform data rebalancing. While this could be viewed as a limitation, it reflects real-world scenarios where models must adapt with limited training data \citep{bassignana-plank-2022-crossre}. Additionally, we evaluate replicated SOTA systems on identical biographical subsets as our models rather than full test sets to ensure controlled comparison with our models. Finally, our work is limited to examining cross-dataset generalisation across three general-purpose datasets due to limited relation type overlap across RE datasets.

\bibliography{anthology,custom}

\appendix
\label{sec:appendix}
\clearpage

\section{Dataset Statistics: Class Distribution}
\label{sec:dataset_statistics}

\begin{table}[!htp]
\centering
\tiny 
\sisetup{
    detect-all,
    group-digits=true,
    group-separator={,},
    input-ignore={,},
    output-decimal-marker={.},
    table-number-alignment=center,
}
\NewDocumentCommand{\B}{}{\fontseries{b}\selectfont}
\begin{tabular}{
  @{}
  l
  S[table-format=6.0, table-space-text-post={*}]
  @{}
}
\toprule
\textbf{Relation} & {\textbf{\# of Samples}} \\
\midrule
Other & 10,000 \\
birthdate & 2,914 \\
bplace\_name & 2,845 \\
dplace\_name & 1,138 \\
occupation & 1,105 \\
deathdate & 1,011 \\
parent & 394 \\
educatedAt & 339 \\
child & 136 \\
sibling & 118 \\
\midrule
Positive Samples & 10,000 \\
Negative Samples & 10,000 \\
All & 20,000  \\
\bottomrule
\end{tabular}
\caption{Balanced Biographical Dataset.}
\label{balanced-biographical-table}

\end{table}

\begin{table}[ht]
\centering
\tiny 
\sisetup{
    detect-all,
    group-digits=true,
    group-separator={,},
    input-ignore={,},
    output-decimal-marker={.},
    table-number-alignment=center,
}
\NewDocumentCommand{\B}{}{\fontseries{b}\selectfont}
\begin{tabular}{
  @{}
  l
  S[table-format=6.0, table-space-text-post={*}]
  @{}
}
\toprule
\textbf{Relation} & {\textbf{\# of Samples}} \\
\midrule
no\_relation                          & 14192 \\
per:title                             & 3805  \\
per:employee\_of                       & 2104 \\
per:age                                & 818  \\
per:countries\_of\_residence           & 695  \\
per:cities\_of\_residence              & 596  \\
per:origin                             & 652  \\
per:stateorprovinces\_of\_residence    & 444  \\
per:spouse                             & 463  \\
per:date\_of\_death                    & 343  \\
per:children                           & 347  \\
per:cause\_of\_death                   & 318  \\
per:parents                            & 282  \\
per:charges                            & 270  \\
per:other\_family                      & 241  \\
per:siblings                           & 238  \\
per:schools\_attended                  & 219  \\
per:city\_of\_death                    & 204  \\
per:religion                           & 145  \\
per:alternate\_names                   & 132  \\
per:city\_of\_birth                    & 107  \\
per:stateorprovince\_of\_death         & 100  \\
per:date\_of\_birth                    & 99   \\
per:stateorprovince\_of\_birth         & 77   \\
per:country\_of\_death                 & 57   \\
per:country\_of\_birth                 & 45   \\
\midrule
Positive Samples & 12801 \\
Negative Samples & 14192 \\ 
All & 26993  \\ 
\bottomrule
\end{tabular}
\caption{Balanced TACRED-RE Subset with Biographical Relations (26 relations).}
\label{tacred-re-table}
\end{table}




\begin{table}[!htp]
\centering
\tiny 
\sisetup{
    detect-all,
    group-digits=true,
    group-separator={,},
    input-ignore={,},
    output-decimal-marker={.},
    table-number-alignment=center,
}
\begin{tabular}{lr}
\toprule
\textbf{Relation} & \textbf{\# of Samples} \\
\midrule
None & 5068 \\ 
/people/person/nationality & 2160 \\
/people/person/place\_lived & 2016 \\
/people/person/place\_of\_birth & 437 \\
/people/deceased\_person/place\_of\_death & 284 \\
/people/person/children & 147 \\
/people/person/religion & 24 \\
\midrule
Positive Samples & 5068 \\
Negative Samples & 5068 \\
All & 10,136 \\
\bottomrule
\end{tabular}
\caption{Balanced NYT Subset with Biographical Relations after removal of instances with multiple labels (7 relations).}
\label{nyt-table}
\end{table}

\section{Relation Type Overlap}

\begin{table}[htp]
\centering
\tiny 
\begin{tabular}{ll}
\toprule
\textbf{NYT} & \textbf{Biographical} \\
\midrule
\texttt{/people/person/children} & \texttt{child} \\
\texttt{/people/person/place\_of\_birth} & \texttt{bplace\_name} \\
\texttt{/people/deceased\_person/place\_of\_death} & \texttt{dplace\_name} \\
\texttt{None} & \texttt{Other} \\
\bottomrule
\end{tabular}
\caption{NYT/Biographical Relation Overlap.}
\label{tab:nyt_bio_mapping}
\end{table}

\begin{table}[htp]
\centering
\tiny 
\setlength{\tabcolsep}{6pt} 
\renewcommand{\arraystretch}{1.2} 
\begin{tabular}{p{3.8cm} p{3cm}} 
\toprule
\textbf{NYT} & \textbf{TACRED-RE} \\
\midrule
None & no\_relation \\
\midrule
/people/person/children & per:children \\
\midrule
/people/person/religion & per:religion \\
\midrule
/people/person/place\_lived & per:stateorprovinces\_of\_residence \\
                          & per:countries\_of\_residence \\
                          & per:cities\_of\_residence \\ 
\midrule
/people/person/place\_of\_birth & per:stateorprovince\_of\_birth \\
                               & per:country\_of\_birth \\
                               & per:city\_of\_birth \\
\midrule
/people/deceased\_person/place\_of\_death & per:stateorprovince\_of\_death \\
                                          & per:country\_of\_death \\
                                          & per:city\_of\_death \\   
\bottomrule
\end{tabular}
\caption{NYT/TACRED-RE Relation Overlap.}
\label{tab:overlap_nyt_tacred}
\end{table}

\begin{table}[htp]
\centering
\tiny 
\begin{tabular}{ll}
\toprule
\textbf{Biographical} & \textbf{TACRED-RE} \\
\midrule
\multirow{3}{*}{\texttt{bplace\_name}} & \texttt{per:stateorprovince\_of\_birth} \\
  & \texttt{per:country\_of\_birth} \\
  & \texttt{per:city\_of\_birth} \\
\midrule
\texttt{birthdate} & \texttt{per:date\_of\_birth} \\
\midrule
\texttt{deathdate} & \texttt{per:date\_of\_death} \\
\midrule
\texttt{parent} & \texttt{per:parents} \\
\midrule
\texttt{educatedAt} & \texttt{per:schools\_attended} \\
\midrule
\multirow{3}{*}{\texttt{dplace\_name}} & \texttt{per:stateorprovince\_of\_death}\\
  & \texttt{per:country\_of\_death} \\
  & \texttt{per:city\_of\_death} \\
\midrule
\texttt{sibling} & \texttt{per:siblings} \\
\midrule
\texttt{child} & \texttt{per:children} \\
\midrule
\texttt{Other} & \texttt{no\_relation} \\

\bottomrule
\end{tabular}
\caption{Biographical/TACRED-RE Relation Overlap.}
\label{tab:bio_tacred_overlap}
\end{table}

\begin{table}[htp]
\centering
\tiny 
\setlength{\tabcolsep}{3pt} 
\renewcommand{\arraystretch}{0.9} 
\begin{tabular}{p{1.1cm} p{2.9cm} p{2.9cm}} 
\toprule
\textbf{Biographical} & \textbf{TACRED-RE} & \textbf{NYT} \\
\midrule
\texttt{child} & \texttt{per:children} & \texttt{/people/person/children} \\
\midrule
\multirow{3}{*}{\texttt{bplace\_name}} & \texttt{per:stateorprovince\_of\_birth} & \multirow{3}{*}{\texttt{/people/person/place\_of\_birth}} \\
  & \texttt{per:country\_of\_birth} &  \\
  & \texttt{per:city\_of\_birth} &  \\
\midrule
\multirow{3}{*}{\texttt{dplace\_name}} & \texttt{per:stateorprovince\_of\_death} & \multirow{3}{*}{\parbox{3cm}{\texttt{/people/deceased\_person/\\place\_of\_death}}} \\
  & \texttt{per:country\_of\_death} &  \\
  & \texttt{per:city\_of\_death} &  \\
\midrule
\texttt{Other} & \texttt{no\_relation} & \texttt{None} \\
\bottomrule
\end{tabular}
\caption{Biographical/TACRED-RE/NYT Overlap.}
\label{tab:overlap_biographical_tacred_nyt}
\end{table}

\begin{figure}[ht]
    \centering
    \includegraphics[width=\linewidth]{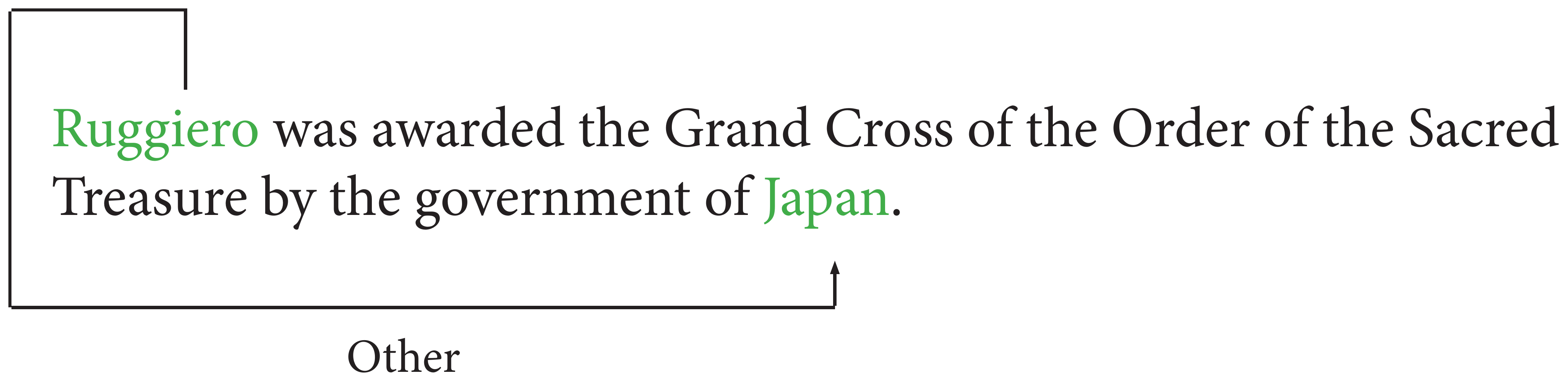}
    \caption{Example for `Other' relation from the Biographical dataset (sample ID `mS7/1269356').}
    \label{fig:bio_example_other}
\end{figure}

\newpage
\section{Implementation Details}\label{sec:implementation_details}
\subsection{Data Formatting Details}\label{sec:data_processing}

TACRED-RE's fine-grained relations (e.g., "per:city\_of\_birth" and "per:country\_of\_birth") were mapped to broader categories (e.g., "place\_of\_birth") used in NYT and Biographical datasets, as shown in Tables~\ref{tab:overlap_nyt_tacred} and~\ref{tab:bio_tacred_overlap}. Cross-dataset results are reported using NYT label names (Table~\ref{tab:cross-dataset-nyt-preds-per-class-results}).

\subsection{Model Implementation}\label{sec:model_implementation}
 We fine-tuned DeBERTa-v3-large\footnote{\url{https://huggingface.co/microsoft/deberta-v3-large}} for 10 epochs employing early stopping. Following prior work using encoder-based models for RE \citep{baldini-soares-etal-2019-matching,10.1145/3477495.3531742}, we extended the \texttt{DeBERTa-v3-large} tokeniser with entity marker tokens, namely, \texttt{<e1>} and \texttt{</e1>} for the head entity and \texttt{<e2>} and \texttt{</e2>} for the tail entity, and used the concatenated final hidden states of the entity start tokens, specifically \texttt{<e1>} and \texttt{<e2>}, as input to the classification head. For LLaMA 3.1\footnote{\url{https://huggingface.co/meta-llama/Llama-3.1-8B-Instruct}}, we used LoRA fine-tuning ($r=8$) over three epochs, applying it to attention and feedforward modules. Both models were fine-tuned using HuggingFace's Trainer class. For evaluation of LLaMA 3.1, predictions were considered correct only if they matched ground-truth labels exactly \citep{hendryckstest2021}.

 For prompting, we used \texttt{vanilla} prompting \citep{li-etal-2023-revisiting-large,vatsal2024surveypromptengineeringmethods} and tested several RE-specific prompt designs  \citep{leidinger-etal-2023-language,li-etal-2023-revisiting-large,efeoglu2024relationextractionfinetunedlarge}, given LLaMA's sensitivity to prompt formulation \citep{leidinger-etal-2023-language}. The prompt~\ref{lst:llama_prompt} performed best and was used across all datasets with adapted label sets. Further prompt optimisation techniques were not considered, as they were beyond the scope of this paper. Hyperparameter settings for all experiments are detailed in Table~\ref{tab:hyperparams_combined}.

  \begin{lstlisting}[caption={System prompt for LLaMA 3.1 8B zero-shot, few-shot, and fine-tuning experiments, shown here with NYT relation inventory.}, label={lst:llama_prompt}, basicstyle=\ttfamily\scriptsize, breaklines=true, frame=single, captionpos=b]
system_message = {
    "role": "system",
    "content": (
            "You are an intelligent assistant specializing in identifying relations between entities in a sentence. "
            "Question: What is the relation between two tagged entities <e1>entity1</e1> and <e2>entity2</e2> in the following sentence? "
            "Choose one relation from the list: "
            "['/people/person/children', '/people/person/nationality', '/people/person/place_lived',"
             "'/people/person/place_of_birth', '/people/deceased_person/place_of_death', '/people/person/religion',"
             "'None']. "
            "Rules: Select exactly one relation from the list. If none of the listed relations apply, select 'None'. "
             "Output must strictly follow this format: <relation_type>. Provide no additional text or explanation."
    )
}
\end{lstlisting}

Due to Biographical's ambiguous `Other' class (Section~\ref{sec:data}), we use it only for evaluation in the main paper (Table~\ref{tab:results_bio_overlap_macro}, Section~\ref{sec:results}) and add supplementary cross-dataset experiments with it in the Appendix Section~\ref{sec:cross_dataset_bio}.

All experiments with DeBERTa-v3-large were run on a single NVIDIA\textsuperscript{\textregistered} TITAN RTX 24GB GPU; all experiments with LLaMA-3.1-8B-Instruct were run on a single NVIDIA\textsuperscript{\textregistered} A100 80GB GPU. All experiments were performed with a fixed random seed for reproducibility.

\subsection{Replication Study}\label{sec:replication_sota}
For our replication of SOTA models for NYT \citep{orlando-etal-2024-relik}, TACRED-RE \citep{zhou-chen-2022-improved}, and Biographical \citep{10.1145/3477495.3531742}, we either used the provided trained model, implementation, or closely followed the implementation details provided in the corresponding paper. Thus, all models were trained on full training data (i.e., train and validation) (NYT: $\sim$266k, TACRED-RE: $\sim$107k, Biographical: $\sim$346k instances), while our models used only downsampled biographical subsets (NYT: $\sim$10k, TACRED-RE: $\sim$26k, Biographical: $\sim$20k instances), as reflected in Tables~\ref{balanced-biographical-table}, ~\ref{tacred-re-table}, and~\ref{nyt-table}. The evaluation was performed on the biographical test sets used in our experiments.

For NYT \citep{orlando-etal-2024-relik}, the retriever-reader \texttt{ReLiK Large} model is accessible via HuggingFace\footnote{\url{https://huggingface.co/sapienzanlp/relik-relation-extraction-nyt-large}} ecosystem and it was used for evaluation on the biographical test set. Detailed per-class results of the model by \citet{orlando-etal-2024-relik} on this test set along with results using our models are reflected in Table~\ref{nyt-results}. While \citet{orlando-etal-2024-relik} reports micro F1 of 0.95 on the full test set, ReLiK-Large achieves 0.90 micro F1 on the biographical subset, underperforming our fine-tuned LLaMA model by 4 micro F1 points, as reflected in Table~\ref{nyt-results}.

For TACRED-RE, we replicated the SOTA model by \citet{zhou-chen-2022-improved}, RoBERTa-large with typed entity markers that uses entity-specific hidden representations as input for the classification head, using provided code\footnote{\url{https://github.com/wzhouad/RE_improved_baseline}}. Detailed per-class results of the model by \citet{zhou-chen-2022-improved} on this test set along with results using our models are reflected in Table~\ref{tacred-re-results}. On the full test set of TACRED-RE \citep{alt-etal-2020-tacred}, \citet{zhou-chen-2022-improved} report micro F1 of 0.83, whereas it achieves a micro F1 of 0.87 on the biographical subset trailing our fine-tuned LLaMA model by 2 micro F1 points.

For Biographical, we replicated the SOTA model \citet{10.1145/3477495.3531742}, who used BERT Base with entity markers and entity-specific hidden representations for classification—the same representation strategy we employed in our DeBERTa experiments. Since neither the trained BERT Base model nor the train-validation-test split used by \citet{10.1145/3477495.3531742} were made publicly available, the results cannot be considered fully replicable. We followed all implementation details provided by \citet{10.1145/3477495.3531742} to replicate their model. On the full test set of the \texttt{normal} version of Biographical, as designated by the authors, \citet{10.1145/3477495.3531742} report macro F1 of 0.76. With our replicated model, we achieve macro F1 of 0.87 and micro F1 of 0.93 on the downsampled test set, outperforming our DeBERTa by 4 macro F1 points and 1 micro F1 point (Table~\ref{tab:results_bio}). This gap is likely attributable to training data disparity—we heavily downsampled Biographical to 20K instances to match NYT and TACRED-RE subset sizes, whereas the SOTA approach used the full 346K instances.

\section{Vocabulary Overlap between Datasets}

\begin{figure}[t]
    \centering
    \includegraphics[width=\linewidth]
    {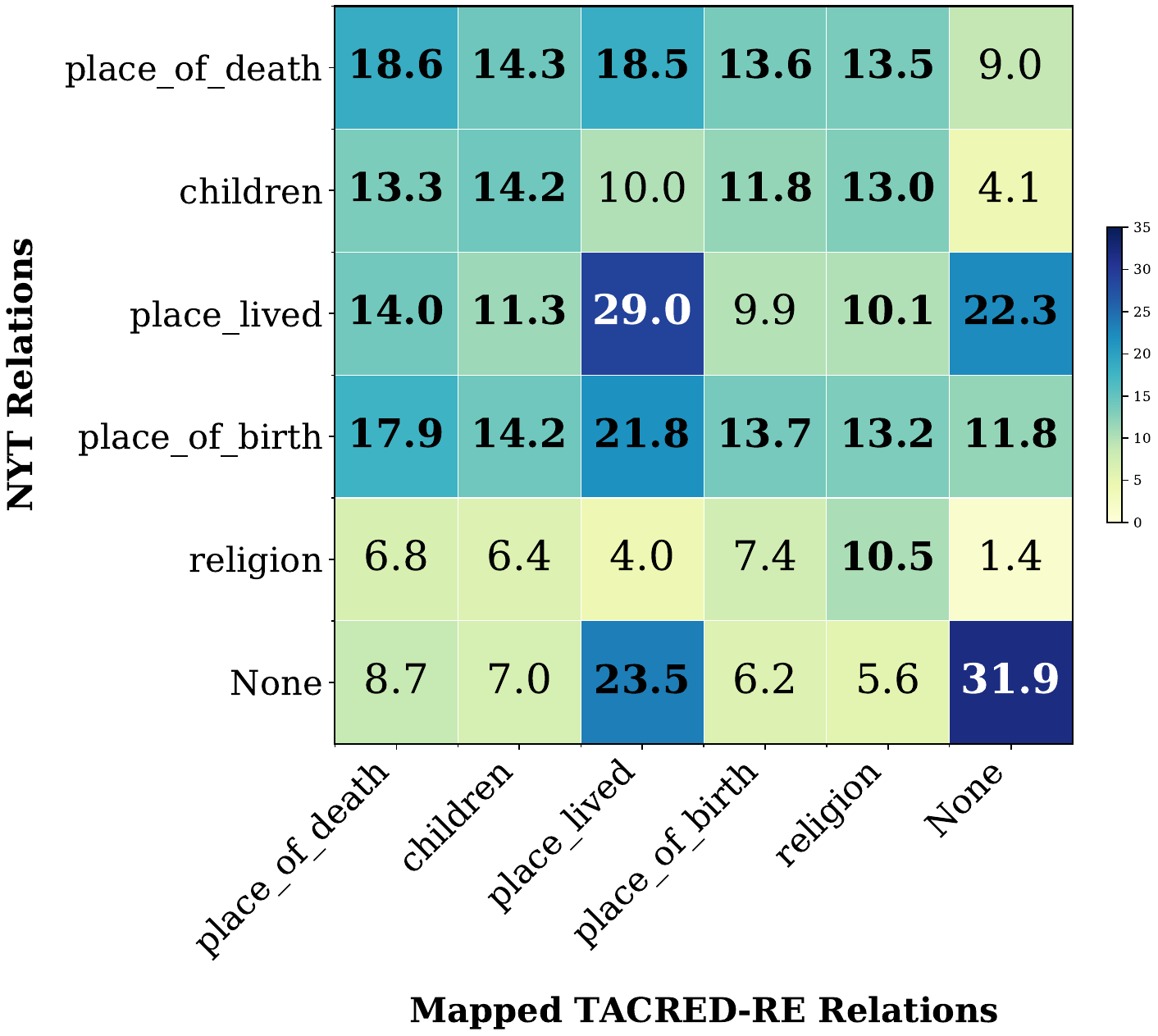}
    \caption{Vocabulary Overlap (\%) per overlapping relation between NYT and TACRED-RE.}
    \label{fig:vocab_overlap_nyt_tacrev}
\end{figure}

\begin{figure}[t]
    \centering
    \includegraphics[width=\linewidth]
    {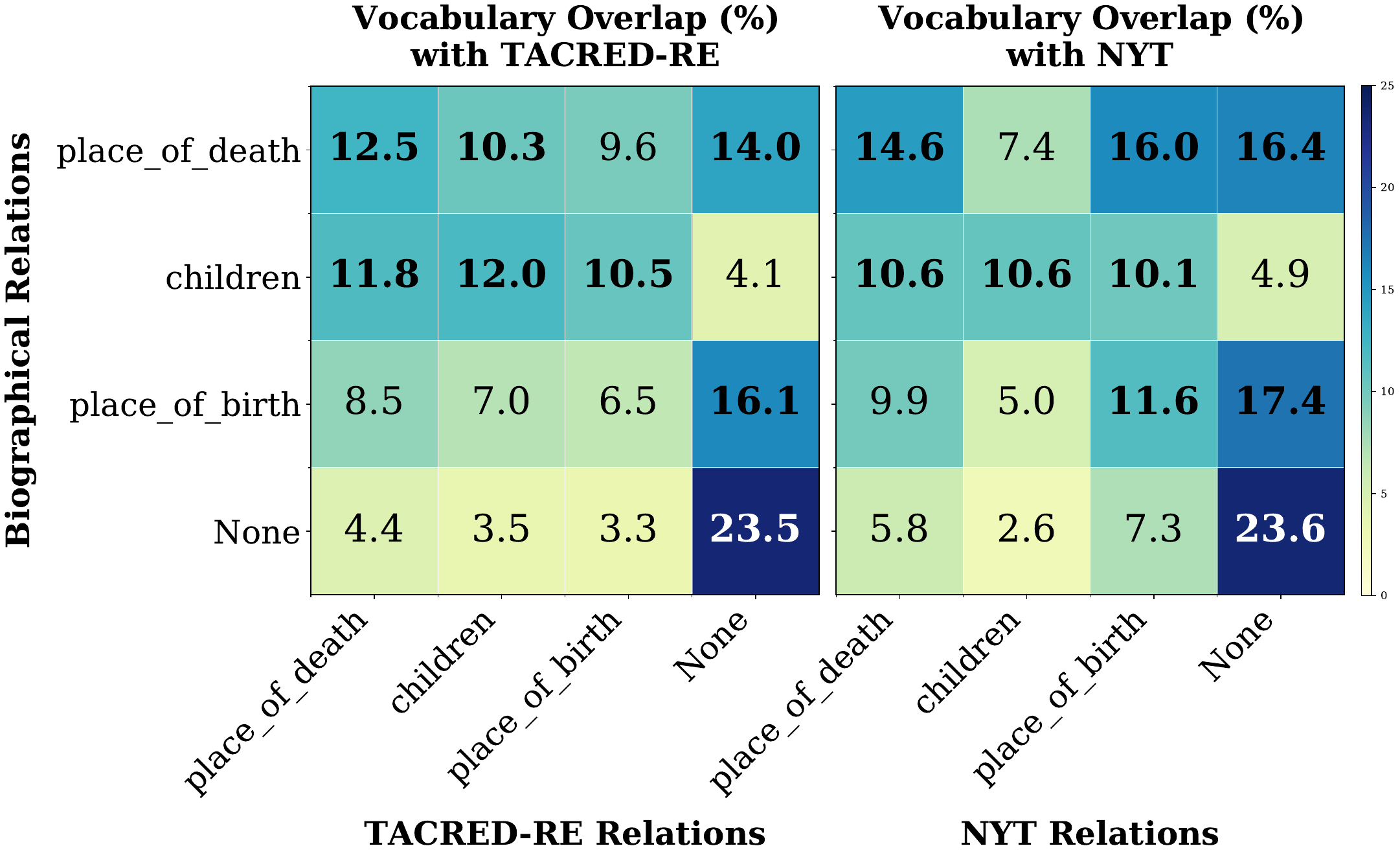}
    \caption{Vocabulary Overlap (\%) Between Biographical and TACRED-RE/NYT Relations following lemmatisation and stop word removal using spaCy's \texttt{en\_core\_web\_trf}.}
    \label{fig:vocab_overlap_bio}
\end{figure}

Figure~\ref{fig:vocab_overlap_nyt_tacrev} depicts vocabulary overlap between NYT and TACRED-RE per overlapping relation. Figure~\ref{fig:vocab_overlap_bio} depicts vocabulary overlap between Biographical and TACRED-RE as well as Biographical and NYT per overlapping relation.

\section{Cross-Dataset Results with Biographical-adapted Models}\label{sec:cross_dataset_bio}

\begin{table*}[t]
\centering
\tiny
\setlength{\tabcolsep}{5pt}
\begin{tabular}{l|l|l|c|c|c|c|c}
\hline
\textbf{Model} & \textbf{Setting} & \textbf{Dataset} & \multicolumn{2}{c|}{\textbf{Intra-Dataset}} & \multicolumn{3}{c}{\textbf{Cross-Dataset}} \\
\cline{4-8}
& & & \textbf{Shared Labels} & \textbf{Dataset Labels} & \textbf{NYT} & \textbf{TACRED-RE} & \textbf{Biographical} \\
\hline
DeBERTa-v3 large 304M & Fine-tuned on & NYT & \textbf{0.83} & 0.83 & -- & 0.28 & 0.48 \\
& & TACRED-RE & 0.73 & 0.57 & 0.55 & -- & 0.62 \\
& & Biographical & \textbf{0.80} & 0.80 & 0.46 & 0.62 & -- \\
\hline
LLaMA 3.1 8B & Fine-tuned on & NYT & 0.82 & 0.82 & -- & 0.43 & 0.30 \\
& & TACRED-RE & \textbf{0.82} & 0.74 & 0.55 & -- & \textbf{0.69} \\
& & Biographical & 0.79 & 0.79 & \textbf{0.65} & 0.65 & -- \\
\hline
LLaMA 3.1 8B & Zero-Shot & NYT & 0.30 & 0.30 & -- & -- & -- \\
& & TACRED-RE & 0.58 & 0.39 & -- & -- & -- \\
& & Biographical & 0.24 & 0.24 & -- & -- & -- \\
\hline
LLaMA 3.1 8B & 5-Shot & NYT & 0.45 $\pm$ 0.04 & 0.45 $\pm$ 0.04 & -- & 0.62 $\pm$ 0.06 & 0.48 $\pm$ 0.04 \\
& & TACRED-RE & 0.70 $\pm$ 0.13 & 0.47 $\pm$ 0.08 & 0.37 $\pm$ 0.03 & -- & 0.51 $\pm$ 0.04 \\
& & Biographical & 0.53 $\pm$ 0.05 & 0.53 $\pm$ 0.05 & 0.32 $\pm$ 0.05 & \textbf{0.68 $\pm$ 0.02} & -- \\
\hline
\end{tabular}
\caption{Macro F1-scores for intra- and cross-dataset predictions on four overlapping relations. Results show both shared and dataset-specific labels, with models adapted on all biographical relations through fine-tuning or ICL. The best intra- and cross-dataset results on full overlap are highlighted in bold.}
\label{tab:results_cross_dataset_4rels}
\end{table*}

We extend our cross-dataset experiments with a cross-dataset evaluation of models adapted on Biographical (Table~\ref{tab:results_cross_dataset_4rels}). 
We compare these cross-dataset experiments with NYT- and TACRED-RE-adapted models evaluated on the same four overlapping relations between the three datasets.

As observed with other datasets, Biographical-adapted models also exhibit performance degradation in cross-dataset scenarios, with NYT and TACRED-RE scores showing drops of 14-17 points. Following patterns observed in Section~\ref{sec:results}, LLaMA fine-tuned on Biographical achieves higher scores than DeBERTa on both NYT and TACRED-RE. 

Notably, Biographical-adapted LLaMA achieves the best cross-dataset results on NYT, even outperforming its TACRED-RE counterpart. However, closer manual analysis of errors suggests that this stems from the Biographical dataset's alignment with NYT's annotation methodology: for example, in ``\texttt{Arne Duncan}, the chief executive of \texttt{Chicago} public schools [...]''\footnote{NYT instance ID: /m/vinci8/data1/riedel/projects/relation\allowbreak/kb\allowbreak/nyt1\allowbreak/docstore\allowbreak/nyt-2005-2006.backup/1653431.xml.pb}, the Biographical-adapted model predicts `place\_of\_birth' (matching NYT's gold label) despite absent textual evidence, while TACRED-RE more appropriately predicts `no\_relation'. Thus, this finding does not necessarily indicate Biographical-adapted models possess better cross-dataset transfer capabilities than models finetuned on the other datasets. 

In few-shot ICL, LLaMA with Biographical demonstrations underperforms its TACRED-RE counterpart on NYT by 5 points, while achieving nearly identical results on TACRED-RE as compared to intra-dataset few-shot ICL results. Biographical-adapted models also achieve their best cross-dataset results on TACRED-RE via few-shot ICL adaptation (0.68 $\pm$ 0.02), which aligns with our hypothesis in Section~\ref{sec:cross_dataset_res} that ICL mitigates noisy training signals (see Section~\ref{sec:performance_analysis} for more details on Biographical annotation noise). 

\begin{table*}[htp]
\section{Results}\label{sec:appendix_results}
\subsection{Intra-Dataset Results}\label{sec:intra-dataset-results}
\vspace{1em}
\centering
\tiny
\begin{tabular}{l|>{\centering\arraybackslash}p{0.5cm}>{\centering\arraybackslash}p{0.5cm}>{\centering\arraybackslash}p{0.5cm}||>{\centering\arraybackslash}p{0.5cm}>{\centering\arraybackslash}p{0.5cm}>{\centering\arraybackslash}p{0.5cm}|>{\centering\arraybackslash}p{0.5cm}>{\centering\arraybackslash}p{0.5cm}>{\centering\arraybackslash}p{0.5cm}|>{\centering\arraybackslash}p{0.5cm}>{\centering\arraybackslash}p{0.5cm}>{\centering\arraybackslash}p{0.5cm}|>{\centering\arraybackslash}p{0.5cm}>{\centering\arraybackslash}p{0.5cm}>{\centering\arraybackslash}p{0.5cm}}
\toprule
\textbf{Model} & \multicolumn{3}{c||}{\textbf{\citet{orlando-etal-2024-relik}}} & \multicolumn{3}{c|}{\textbf{DeBERTa-v3-large 304M}} & \multicolumn{3}{c|}{\textbf{LLaMA-3.1 8B zero-shot}} & \multicolumn{3}{c|}{\textbf{LLaMA-3.1 8B 5-shot}} & \multicolumn{3}{c}{\textbf{LLaMA-3.1 8B fine-tuned}} \\
\midrule
 & \textbf{P} & \textbf{R} & \textbf{F1} & \textbf{P} & \textbf{R} & \textbf{F1} & \textbf{P} & \textbf{R} & \textbf{F1} & \textbf{P} & \textbf{R} & \textbf{F1} & \textbf{P} & \textbf{R} & \textbf{F1} \\
\midrule
place\_of\_death & 0.84 & 0.97 & 0.90 & 0.82 & 0.70 & 0.75 & 0.93 & 0.82 & 0.87 & 0.75 & 0.09 & 0.16 & 0.93 & 0.82 & 0.87 \\
children & 0.48 & 0.93 & 0.63 & 0.87 & 0.93 & 0.90 & 0.87 & 0.93 & 0.90 & 0.77 & 0.71 & 0.74 & 0.87 & 0.93 & 0.90 \\
nationality & 0.94 & 1.00 & 0.97 & 0.98 & 1.00 & 0.99 & 0.99 & 0.99 & 0.99 & 0.96 & 0.10 & 0.18 & 0.99 & 0.99 & 0.99 \\
place\_lived & 0.84 & 0.98 & 0.90 & 0.85 & 0.93 & 0.89 & 0.84 & 0.95 & 0.89 & 0.36 & 0.58 & 0.44 & 0.84 & 0.95 & 0.89 \\
place\_of\_birth & 0.70 & 0.94 & 0.80 & 0.76 & 0.54 & 0.63 & 0.88 & 0.46 & 0.60 & 0.07 & 0.06 & 0.07 & 0.88 & 0.46 & 0.60 \\
religion & 1.00 & 1.00 & 1.00 & 1.00 & 0.60 & 0.75 & 1.00 & 1.00 & 1.00 & 0.83 & 1.00 & 0.91 & 1.00 & 1.00 & 1.00 \\
None & 0.98 & 0.81 & 0.89 & 0.97 & 0.96 & 0.97 & 0.97 & 0.97 & 0.97 & 0.62 & 0.76 & 0.68 & 0.97 & 0.97 & 0.97 \\
\midrule
macro avg & 0.87 & 0.87 & 0.87 & 0.89 & 0.81 & 0.84 & 0.92 & 0.87 & 0.89 & 0.62 & 0.47 & 0.45 & 0.92 & 0.87 & \textbf{0.89} \\
micro avg & 0.90 & 0.90 & 0.90 & 0.93 & 0.93 & 0.93 & 0.30 & 0.22 & 0.25 & 0.52 & 0.52 & 0.52 & 0.94 & 0.94 & \textbf{0.94} \\
weighted avg & 0.90 & 0.90 & 0.90 & 0.93 & 0.93 & 0.93 & 0.94 & 0.94 & 0.94 & 0.63 & 0.52 & 0.47 & 0.94 & 0.94 & \textbf{0.94} \\
\bottomrule
\end{tabular}
\caption{NYT Results. SOTA model by \citet{orlando-etal-2024-relik} used full NYT train/dev sets, while ours were adapted exclusively on the subset with biographical relations.}
\label{nyt-results}
\end{table*}

\begin{table*}[htp]
\centering
\tiny
\begin{tabular}{l|>{\centering\arraybackslash}p{0.5cm}>{\centering\arraybackslash}p{0.5cm}>{\centering\arraybackslash}p{0.5cm}|>{\centering\arraybackslash}p{0.5cm}>{\centering\arraybackslash}p{0.5cm}>{\centering\arraybackslash}p{0.5cm}|>{\centering\arraybackslash}p{0.5cm}>{\centering\arraybackslash}p{0.5cm}>{\centering\arraybackslash}p{0.5cm}|>{\centering\arraybackslash}p{0.5cm}>{\centering\arraybackslash}p{0.5cm}>{\centering\arraybackslash}p{0.5cm}}
\toprule
\textbf{Model} & \multicolumn{3}{c|}{\textbf{DeBERTa-v3-large 304M}} & \multicolumn{3}{c|}{\textbf{LLaMA-3.1 8B zero-shot}} & \multicolumn{3}{c|}{\textbf{LLaMA-3.1 8B 5-shot}} & \multicolumn{3}{c}{\textbf{LLaMA-3.1 8B fine-tuned}} \\
\midrule
 & \textbf{P} & \textbf{R} & \textbf{F1} & \textbf{P} & \textbf{R} & \textbf{F1} & \textbf{P} & \textbf{R} & \textbf{F1} & \textbf{P} & \textbf{R} & \textbf{F1} \\
\midrule
macro avg & 0.67 & 0.64 & 0.64 & 0.49 & 0.38 & 0.34 & 0.37 & 0.30 & 0.29 & 0.76 & 0.71 & \textbf{0.73} \\
micro avg & 0.83 & 0.83 & 0.83 & 0.31 & 0.29 & 0.30 & 0.51 & 0.51 & 0.51 & 0.87 & 0.87 & \textbf{0.87} \\
weighted avg & 0.83 & 0.83 & 0.83 & 0.68 & 0.29 & 0.32 & 0.59 & 0.51 & 0.49 & 0.87 & 0.87 & \textbf{0.87} \\
\bottomrule
\end{tabular}
\caption{TACRED Results.}
\label{tacred-results}
\end{table*}

\begin{table*}[htp]
\centering
\tiny
\begin{tabular}{l|ccc||ccc|ccc|ccc|ccc}
\toprule
\textbf{Model} & \multicolumn{3}{c||}{\textbf{\citet{zhou-chen-2022-improved}}} & \multicolumn{3}{c|}{\textbf{DeBERTa-v3-large 304M}} & \multicolumn{3}{c|}{\textbf{LLaMA-3.1 8B zero-shot}} & \multicolumn{3}{c|}{\textbf{LLaMA-3.1 8B 5-shot}} & \multicolumn{3}{c}{\textbf{LLaMA-3.1 8B fine-tuned}} \\
\midrule
 & \textbf{P} & \textbf{R} & \textbf{F1} & \textbf{P} & \textbf{R} & \textbf{F1} & \textbf{P} & \textbf{R} & \textbf{F1} & \textbf{P} & \textbf{R} & \textbf{F1} & \textbf{P} & \textbf{R} & \textbf{F1} \\
\midrule
no\_relation & 0.90 & 0.95 & 0.92 & 0.97 & 0.80 & 0.88 & 0.81 & 0.01 & 0.02 & 0.80 & 0.61 & 0.69 & 0.97 & 0.87 & 0.92 \\
per:age & 0.96 & 0.96 & 0.96 & 0.92 & 0.99 & 0.96 & 0.97 & 0.32 & 0.48 & 0.93 & 0.65 & 0.77 & 0.95 & 1.00 & 0.97 \\
per:cause\_of\_death & 0.92 & 0.86 & 0.89 & 0.77 & 0.71 & 0.74 & 0.46 & 0.26 & 0.33 & 0.53 & 0.24 & 0.33 & 0.65 & 0.95 & 0.77 \\
per:charges & 0.98 & 0.88 & 0.92 & 0.82 & 0.97 & 0.89 & 0.76 & 0.30 & 0.43 & 0.67 & 0.43 & 0.52 & 0.87 & 0.99 & 0.93 \\
per:children & 0.93 & 0.68 & 0.78 & 0.65 & 0.81 & 0.72 & 0.23 & 0.14 & 0.17 & 0.19 & 0.38 & 0.25 & 0.96 & 0.73 & 0.83 \\
per:cities\_of\_residence & 0.76 & 0.87 & 0.81 & 0.49 & 0.96 & 0.65 & 0.38 & 0.55 & 0.45 & 0.51 & 0.34 & 0.41 & 0.61 & 0.92 & 0.73 \\
per:city\_of\_birth & 1.00 & 0.50 & 0.67 & 0.75 & 0.50 & 0.60 & 0.50 & 0.67 & 0.57 & 0.43 & 0.50 & 0.46 & 1.00 & 0.50 & 0.67 \\
per:city\_of\_death & 0.60 & 0.38 & 0.46 & 0.64 & 0.44 & 0.52 & 0.33 & 0.31 & 0.32 & 0.28 & 0.56 & 0.38 & 0.43 & 0.56 & 0.49 \\
per:countries\_of\_residence & 0.70 & 0.53 & 0.61 & 0.47 & 0.82 & 0.60 & 0.33 & 0.45 & 0.38 & 0.34 & 0.31 & 0.32 & 0.59 & 0.91 & 0.71 \\
per:country\_of\_death & 0.00 & 0.00 & 0.00 & 0.00 & 0.00 & 0.00 & 0.00 & 0.00 & 0.00 & 0.50 & 0.56 & 0.53 & 0.50 & 0.44 & 0.47 \\
per:date\_of\_birth & 1.00 & 1.00 & 1.00 & 0.78 & 1.00 & 0.88 & 0.26 & 0.86 & 0.40 & 0.71 & 0.71 & 0.71 & 0.86 & 0.86 & 0.86 \\
per:date\_of\_death & 0.94 & 0.73 & 0.82 & 0.65 & 0.88 & 0.74 & 0.53 & 0.42 & 0.47 & 0.62 & 0.12 & 0.21 & 0.74 & 0.93 & 0.82 \\
per:employee\_of & 0.93 & 0.81 & 0.87 & 0.79 & 0.88 & 0.83 & 0.10 & 0.97 & 0.19 & 0.17 & 0.73 & 0.27 & 0.86 & 0.89 & 0.88 \\
per:origin & 0.85 & 0.81 & 0.83 & 0.71 & 0.84 & 0.77 & 0.65 & 0.12 & 0.21 & 0.45 & 0.13 & 0.20 & 0.82 & 0.79 & 0.80 \\
per:other\_family & 0.92 & 0.97 & 0.95 & 0.53 & 0.89 & 0.67 & 0.00 & 0.00 & 0.00 & 0.11 & 0.59 & 0.19 & 0.61 & 0.97 & 0.75 \\
\midrule
macro avg & 0.84 & 0.74 & 0.78 & 0.67 & 0.78 & 0.71 & 0.43 & 0.30 & 0.25 & 0.53 & 0.44 & 0.41 & 0.77 & 0.83 & \textbf{0.78} \\
micro avg & 0.90 & 0.83 & 0.87 & 0.85 & 0.85 & 0.85 & 0.20 & 0.14 & 0.17 & 0.53 & 0.51 & 0.52 & 0.89 & 0.89 & \textbf{0.89} \\
weighted avg & 0.90 & 0.90 & 0.90 & 0.88 & 0.85 & 0.85 & 0.70 & 0.14 & 0.11 & 0.72 & 0.51 & 0.54 & 0.91 & 0.89 & \textbf{0.89} \\
\bottomrule
\end{tabular}
\caption{TACRED-RE Results. SOTA model by \citet{zhou-chen-2022-improved} used full TACRED-RE train/dev sets, while ours were adapted exclusively on the subset with biographical relations.}
\label{tacred-re-results}
\end{table*}

\begin{table*}[htp]
\centering
\tiny
\begin{tabular}{l|ccc||ccc|ccc|ccc|ccc}
\toprule
\textbf{Model} & \multicolumn{3}{c||}{\textbf{\citet{10.1145/3477495.3531742}}} & \multicolumn{3}{c|}{\textbf{DeBERTa-v3-large 304M}} & \multicolumn{3}{c|}{\textbf{LLaMA-3.1 8B zero-shot}} & \multicolumn{3}{c|}{\textbf{LLaMA-3.1 8B 5-shot}} & \multicolumn{3}{c}{\textbf{LLaMA-3.1 8B fine-tuned}} \\
\midrule
 & \textbf{P} & \textbf{R} & \textbf{F1} & \textbf{P} & \textbf{R} & \textbf{F1} & \textbf{P} & \textbf{R} & \textbf{F1} & \textbf{P} & \textbf{R} & \textbf{F1} & \textbf{P} & \textbf{R} & \textbf{F1} \\
\midrule
Other        & 0.95 & 0.94 & 0.95 & 0.95 & 0.93 & 0.94 & 0.78 & 0.29 & 0.42 & 0.94 & 0.54 & 0.69 & 0.92 & 0.93 & 0.93 \\
birthdate    & 1.00 & 1.00 & 1.00 & 1.00 & 1.00 & 1.00 & 0.53 & 0.91 & 0.67 & 0.83 & 0.91 & 0.87 & 1.00 & 1.00 & 1.00 \\
bplace\_name & 0.83 & 0.92 & 0.87 & 0.86 & 0.87 & 0.87 & 0.82 & 0.08 & 0.14 & 0.69 & 0.82 & 0.75 & 0.87 & 0.85 & 0.86 \\
child        & 0.67 & 0.67 & 0.67 & 0.38 & 0.56 & 0.45 & 0.02 & 0.56 & 0.05 & 0.00 & 0.00 & 0.00 & 0.57 & 0.44 & 0.50 \\
deathdate    & 1.00 & 0.98 & 0.99 & 1.00 & 0.95 & 0.98 & 0.73 & 0.82 & 0.77 & 0.65 & 0.85 & 0.74 & 0.96 & 0.99 & 0.98 \\
dplace\_name & 0.76 & 0.67 & 0.71 & 0.60 & 0.72 & 0.66 & 0.39 & 0.25 & 0.31 & 0.50 & 0.55 & 0.52 & 0.56 & 0.81 & 0.66 \\
educatedAt   & 0.86 & 0.97 & 0.91 & 0.84 & 0.84 & 0.84 & 0.06 & 1.00 & 0.12 & 0.19 & 1.00 & 0.32 & 0.00 & 0.00 & 0.00 \\
occupation   & 1.00 & 1.00 & 1.00 & 1.00 & 1.00 & 1.00 & 0.83 & 0.41 & 0.55 & 0.77 & 0.94 & 0.85 & 1.00 & 0.98 & 0.99 \\
parent       & 0.83 & 0.77 & 0.80 & 0.77 & 0.82 & 0.79 & 0.38 & 0.59 & 0.46 & 0.26 & 0.84 & 0.39 & 1.00 & 0.68 & 0.81 \\
sibling      & 0.85 & 0.73 & 0.79 & 0.83 & 0.67 & 0.74 & 0.10 & 0.13 & 0.11 & 0.42 & 0.33 & 0.37 & 1.00 & 0.67 & 0.80 \\
\midrule
macro avg    & 0.87 & 0.86 & 0.87 & 0.82 & 0.84 & \textbf{0.83} & 0.46 & 0.50 & 0.36 & 0.53 & 0.68 & 0.55 & 0.79 & 0.74 & 0.75 \\
micro avg    & 0.93 & 0.93 & 0.93 & 0.92 & 0.92 & \textbf{0.92} & 0.41 & 0.40 & 0.41 & 0.69 & 0.69 & 0.69 & 0.91 & 0.91 & 0.91 \\
weighted avg & 0.93 & 0.93 & 0.93 & 0.92 & 0.92 & \textbf{0.92} & 0.70 & 0.40 & 0.43 & 0.80 & 0.69 & 0.71 & 0.90 & 0.91 & 0.90 \\
\bottomrule
\end{tabular}
\caption{Biographical Results. The model by \citet{10.1145/3477495.3531742} used full Biographical train/dev sets, while ours were adapted on its downsampled subset matching NYT and TACRED-RE biographical subset sizes.}
\label{tab:results_bio}
\end{table*}

\begin{table*}[ht]
\subsection{Cross-Dataset Results with Models Fine-tuned on Overlap}
\vspace{1em}
\centering
\tiny
\setlength{\tabcolsep}{5pt}
\begin{tabular}{l|l|l|c|c|c|c}
\hline
\textbf{Model} & \textbf{Setting} & \textbf{Dataset} & \multicolumn{2}{c|}{\textbf{Intra-Dataset}} & \multicolumn{2}{c}{\textbf{Cross-Dataset}} \\
\hline
 &  & & \textbf{Shared Labels} & \textbf{Dataset Labels} & \textbf{NYT} & \textbf{TACRED-RE} \\
\hline
DeBERTa-v3-large 304M & Fine-tuned on & NYT & 0.79 & 0.79 & -- & 0.27 \\
 &  & TACRED-RE & 0.73 & 0.64 & 0.49 & -- \\
\hline
LLaMA 3.1 8B & Fine-tuned on & NYT & 0.83 & 0.83 & -- & 0.45 \\
 &  & TACRED-RE & 0.79 & 0.76 & 0.58 & -- \\
\hline
\hline
 & Shot Setting &  &  &  &  &  \\
 \hline
LLaMA 3.1 8B & Zero-Shot & NYT & 0.35 & 0.35 & -- & -- \\
 &  & TACRED-RE & 0.64 & 0.54 & -- & -- \\
\hline
LLaMA 3.1 8B & 5-Shot & NYT & 0.46 $\pm$ 0.04 & 0.46 $\pm$ 0.04 & - & 0.50 $\pm$ 0.06 \\
 &  & TACRED-RE & 0.59 $\pm$ 0.03 & 0.45 $\pm$ 0.05 & 0.44 $\pm$ 0.04 & - \\
\hline
\end{tabular}
\caption{Macro F1-scores for intra- and cross-dataset predictions for six overlapping relations. Results are reported for shared and dataset-specific labels in both fine-tuned and shot settings; only the six overlapping relations are used for adaptation.}
\label{tab:results_tacrev_nyt_6rel}
\end{table*}

\begin{table*}[htp]
\subsection{Cross-Dataset Per-Class Results}
\vspace{1em}
\centering
\tiny
\begin{tabular}{l|ccc|ccc|ccc}
\toprule
\textbf{Model} & \multicolumn{3}{c|}{\textbf{DeBERTa-v3-large 304M}} & \multicolumn{3}{c|}{\textbf{LLaMA-3.1 8B 5-shot}} & \multicolumn{3}{c}{\textbf{LLaMA-3.1 8B fine-tuned}} \\
\midrule
 & \textbf{P} & \textbf{R} & \textbf{F1} & \textbf{P} & \textbf{R} & \textbf{F1} & \textbf{P} & \textbf{R} & \textbf{F1} \\
\midrule
/people/deceased\_person/place\_of\_death & 0.75 & 0.36 & 0.49 & 0.54 $\pm$ 0.07 & 0.39 $\pm$ 0.11 & 0.45 $\pm$ 0.08 & 0.71 & 0.45 & 0.56 \\
/people/person/children & 0.75 & 0.64 & 0.69 & 0.47 $\pm$ 0.05 & 0.29 $\pm$ 0.10 & 0.34 $\pm$ 0.08 & 0.69 & 0.64 & 0.67 \\
/people/person/place\_lived & 0.62 & 0.90 & 0.73 & 0.27 $\pm$ 0.06 & 0.22 $\pm$ 0.15 & 0.23 $\pm$ 0.11 & 0.63 & 0.86 & 0.73 \\
/people/person/place\_of\_birth & 0.20 & 0.04 & 0.07 & 0.12 $\pm$ 0.05 & 0.07 $\pm$ 0.04 & 0.09 $\pm$ 0.04 & 0.17 & 0.02 & 0.04 \\
/people/person/religion & 1.00 & 0.20 & 0.33 & 0.64 $\pm$ 0.03 & 0.72 $\pm$ 0.23 & 0.66 $\pm$ 0.11 & 1.00 & 0.80 & 0.89 \\
None & 0.94 & 0.78 & 0.85 & 0.84 $\pm$ 0.03 & 0.46 $\pm$ 0.07 & 0.59 $\pm$ 0.06 & 0.90 & 0.82 & 0.86 \\
\midrule
macro avg & 0.71 & 0.49 & 0.53 & 0.48 $\pm$ 0.03 & 0.36 $\pm$ 0.04 & 0.39 $\pm$ 0.02 & 0.68 & 0.60 & \textbf{0.62} \\
micro avg & 0.79 & 0.74 & 0.77 & 0.59 $\pm$ 0.07 & 0.37 $\pm$ 0.03 & 0.45 $\pm$ 0.04 & 0.79 & 0.76 & \textbf{0.78} \\
weighted avg & 0.80 & 0.74 & 0.75 & 0.62 $\pm$ 0.04 & 0.37 $\pm$ 0.03 & 0.45 $\pm$ 0.03 & 0.78 & 0.76 & \textbf{0.76} \\
\bottomrule
\end{tabular}
\caption{Adapted on TACRED-RE with all biographical relations present in TACRED-RE. Evaluations on NYT. The labels, although borrowed from NYT dataset, reflect the shared labels between NYT and TACRED-RE. More fine-grained TACRED-RE were mapped to broader shared labels to enable cross-dataset evaluation comparison. The best results on full overlap are highlighted in bold.}
\label{tab:cross-dataset-nyt-preds-per-class-results}
\end{table*}

\begin{table*}[htp]
\centering
\tiny
\begin{tabular}{l|ccc|ccc|ccc}
\toprule
\textbf{Model} & \multicolumn{3}{c|}{\textbf{DeBERTa-v3-large 304M}} & \multicolumn{3}{c|}{\textbf{LLaMA-3.1 8B 5-shot}} & \multicolumn{3}{c}{\textbf{LLaMA-3.1 8B fine-tuned}} \\
\midrule
 & \textbf{P} & \textbf{R} & \textbf{F1} & \textbf{P} & \textbf{R} & \textbf{F1} & \textbf{P} & \textbf{R} & \textbf{F1} \\
\midrule
/people/deceased\_person/place\_of\_death & 0.27 & 0.17 & 0.21 & 0.66 $\pm$ 0.11 & 0.33 $\pm$ 0.20 & 0.40 $\pm$ 0.17 & 0.62 & 0.23 & 0.33 \\
/people/person/children & 0.40 & 0.05 & 0.10 & 0.49 $\pm$ 0.15 & 0.48 $\pm$ 0.20 & 0.47 $\pm$ 0.17 & 1.00 & 0.14 & 0.24 \\
/people/person/place\_lived & 0.56 & 0.25 & 0.34 & 0.65 $\pm$ 0.09 & 0.38 $\pm$ 0.06 & 0.48 $\pm$ 0.05 & 0.71 & 0.31 & 0.43 \\
/people/person/place\_of\_birth & 0.00 & 0.00 & 0.00 & 0.18 $\pm$ 0.10 & 0.88 $\pm$ 0.26 & 0.28 $\pm$ 0.11 & 0.38 & 0.25 & 0.30 \\
/people/person/religion & 1.00 & 0.03 & 0.05 & 0.86 $\pm$ 0.04 & 0.64 $\pm$ 0.17 & 0.72 $\pm$ 0.11 & 0.94 & 0.40 & 0.56 \\
None & 0.80 & 0.91 & 0.85 & 0.85 $\pm$ 0.04 & 0.78 $\pm$ 0.12 & 0.81 $\pm$ 0.05 & 0.79 & 0.95 & 0.86 \\
\midrule
macro avg & 0.51 & 0.23 & 0.26 & 0.62 $\pm$ 0.05 & 0.58 $\pm$ 0.06 & \textbf{0.52 $\pm$ 0.06} & 0.74 & 0.38 & 0.45 \\
micro avg & 0.73 & 0.69 & \textbf{0.71} & 0.75 $\pm$ 0.04 & 0.66 $\pm$ 0.09 & \textbf{0.70 $\pm$ 0.06} & 0.74 & 0.74 & \textbf{0.74} \\
weighted avg & 0.72 & 0.69 & 0.67 & 0.79 $\pm$ 0.01 & 0.66 $\pm$ 0.09 & \textbf{0.71 $\pm$ 0.05} & 0.77 & 0.74 & \textbf{0.72} \\
\bottomrule
\end{tabular}
\caption{Adapted on NYT with all biographical relations present in NYT. Evaluations on TACRED-RE. The labels, although borrowed from NYT dataset, reflect the shared labels between NYT and TACRED-RE. More fine-grained TACRED-RE were mapped to broader shared labels to enable cross-dataset evaluation comparison. The best results on full overlap are highlighted in bold.}
\label{tab:cross-dataset-tacrev-preds-per-class-results}
\end{table*}

\begin{table*}[htp]
\centering
\tiny
\begin{tabular}{l|ccc|ccc|ccc}
\toprule
\textbf{Model} & \multicolumn{3}{c|}{\textbf{DeBERTa-v3-large 304M}} & \multicolumn{3}{c|}{\textbf{LLaMA-3.1 8B 5-shot}} & \multicolumn{3}{c}{\textbf{LLaMA-3.1 8B fine-tuned}} \\
\midrule
 & \textbf{P} & \textbf{R} & \textbf{F1} & \textbf{P} & \textbf{R} & \textbf{F1} & \textbf{P} & \textbf{R} & \textbf{F1} \\
\midrule
None & 0.93 & 0.64 & 0.75 & 0.87 $\pm$ 0.10 & 0.49 $\pm$ 0.09 & 0.62 $\pm$ 0.04 & 0.90 & 0.68 & 0.78 \\
/people/person/place\_of\_birth & 0.89 & 0.73 & 0.80 & 0.84 $\pm$ 0.02 & 0.70 $\pm$ 0.08 & 0.76 $\pm$ 0.05 & 0.90 & 0.75 & 0.82 \\
/people/person/children & 0.44 & 0.44 & 0.44 & 0.14 $\pm$ 0.14 & 0.13 $\pm$ 0.14 & 0.14 $\pm$ 0.14 & 0.71 & 0.56 & 0.63 \\
/people/deceased\_person/place\_of\_death & 0.95 & 0.33 & 0.49 & 0.87 $\pm$ 0.05 & 0.36 $\pm$ 0.05 & 0.51 $\pm$ 0.05 & 0.95 & 0.38 & 0.54 \\
\midrule
macro avg & 0.80 & 0.54 & 0.62 & 0.68 $\pm$ 0.06 & 0.42 $\pm$ 0.04 & 0.51 $\pm$ 0.04 & 0.87 & 0.59 & \textbf{0.69} \\
micro avg & 0.91 & 0.63 & 0.74 & 0.84 $\pm$ 0.05 & 0.55 $\pm$ 0.01 & 0.66 $\pm$ 0.01 & 0.90 & 0.67 & \textbf{0.77} \\
weighted avg & 0.91 & 0.63 & 0.74 & 0.85 $\pm$ 0.06 & 0.55 $\pm$ 0.01 & 0.65 $\pm$ 0.01 & 0.90 & 0.67 & \textbf{0.76} \\
\bottomrule
\end{tabular}
\caption{Adapted on TACRED-RE with all biographical relations present in TACRED-RE. Evaluations on Biographical with four biographical relations (full overlap between three datasets). The labels, although borrowed from NYT dataset, reflect the shared labels between NYT, TACRED-RE, and Biographical. The best results on full overlap are highlighted in bold.}
\label{tab:bio-per-class-finetuned-tacrev}
\end{table*}

\begin{table*}[htp]
\centering
\tiny
\begin{tabular}{l|ccc|ccc|ccc}
\toprule
\textbf{Model} & \multicolumn{3}{c|}{\textbf{DeBERTa-v3-large 304M}} & \multicolumn{3}{c|}{\textbf{LLaMA-3.1 8B 5-shot}} & \multicolumn{3}{c}{\textbf{LLaMA-3.1 8B fine-tuned}} \\
\midrule
 & \textbf{P} & \textbf{R} & \textbf{F1} & \textbf{P} & \textbf{R} & \textbf{F1} & \textbf{P} & \textbf{R} & \textbf{F1} \\
\midrule
/people/person/place\_of\_birth & 0.75 & 0.28 & 0.41 & 0.82 $\pm$ 0.02 & 0.71 $\pm$ 0.14 & 0.75 $\pm$ 0.09 & 0.92 & 0.14 & 0.25 \\
/people/person/children & 0.80 & 0.44 & 0.57 & 0.19 $\pm$ 0.06 & 0.51 $\pm$ 0.06 & 0.27 $\pm$ 0.07 & 0.00 & 0.00 & 0.00 \\
/people/deceased\_person/place\_of\_death & 0.72 & 0.12 & 0.21 & 0.76 $\pm$ 0.09 & 0.11 $\pm$ 0.08 & 0.19 $\pm$ 0.12 & 0.74 & 0.13 & 0.22 \\
None & 0.63 & 0.89 & 0.73 & 0.85 $\pm$ 0.09 & 0.62 $\pm$ 0.09 & 0.71 $\pm$ 0.06 & 0.57 & 0.96 & 0.72 \\
\midrule
macro avg & 0.73 & 0.43 & 0.48 & 0.65 $\pm$ 0.02 & 0.49 $\pm$ 0.04 & \textbf{0.48 $\pm$ 0.04} & 0.56 & 0.31 & 0.30 \\
micro avg & 0.65 & 0.57 & 0.60 & 0.59 $\pm$ 0.05 & 0.59 $\pm$ 0.05 & \textbf{0.59 $\pm$ 0.05} & 0.55 & 0.55 & 0.55 \\
weighted avg & 0.69 & 0.57 & 0.55 & 0.82 $\pm$ 0.04 & 0.59 $\pm$ 0.05 & \textbf{0.66 $\pm$ 0.04} & 0.71 & 0.55 & 0.48 \\
\bottomrule
\end{tabular}
\caption{Adapted on NYT with all biographical relations present in NYT. Evaluations on Biographical with four biographical relations (full overlap between three datasets). The labels, although borrowed from the NYT dataset, reflect the shared labels between NYT, TACRED-RE, and Biographical. The best results on full overlap are highlighted in bold.}
\label{tab:bio-per-class-finetuned-nyt}
\end{table*}

\begin{table*}[htbp]
\centering
\scriptsize
\begin{tabular}{l|l|cccc}
\toprule
Setting & Parameter & \makecell{DeBERTa-v3-\\large Fine-tuned} & \makecell{LLaMA 3.1 8B\\Zero-Shot} & \makecell{LLaMA 3.1 8B\\Five-Shot} & \makecell{LLaMA 3.1 8B\\Fine-tuned} \\
\midrule
\multirow{10}{*}{Common} 
 & \# of Epochs & 10 & -- & -- & 3 \\
 & seed & 42 & 42 & 42 & 42 \\
 & Loss & Cross-Entropy Loss & -- & -- & Cross-Entropy Loss \\
 & Optimiser & AdamW & -- & -- & AdamW \\
 & Batch Size & 8 & -- & -- & 4 \\
 & Gradient Accumulation & 4 & -- & -- & -- \\
 & Early Stopping Patience & 2 & -- & -- & 2 \\
 & Temperature & -- & 0.1 & 0.1 & -- \\
 & Nucleus Sampling & -- & 0.9 & 0.9 & -- \\
 & Lora Settings$^{\dagger}$ & -- & -- & -- & 8/32/0.1 \\
& Train-dev-test split & 70-20-10 & -- & -- & 70-20-10 \\
\midrule
\multirow{3}{*}{TACRED-RE} 
 & Learning Rate & \makecell{$5 \times 10^{-6}$ /\\$5 \times 10^{-5}$} & -- & -- & $5 \times 10^{-5}$ \\
 & Max Length & -- & -- & -- & 800/384 \\
 & Max New Tokens & -- & 40 & 256 & -- \\
 & Cross-Validation & --/5-fold & -- & -- & -- \\
\midrule
\multirow{3}{*}{NYT}
 & Learning Rate & $5 \times 10^{-6}$ & -- & -- & $1 \times 10^{-4}$ \\
 & Max Length & -- & -- & -- & 384 \\
 & Max New Tokens & -- & 256 & 256 & -- \\
 & Cross-Validation & --/5-fold & -- & -- & -- \\
\midrule
\multirow{3}{*}{Biographical}
 & Learning Rate & $5 \times 10^{-6}$ & -- & -- & $1 \times 10^{-4}$ \\
 & Max Length & -- & -- & -- & 384 \\
 & Max New Tokens & -- & 40 & 256 & -- \\
\bottomrule
\end{tabular}
\caption{Hyperparameter settings across datasets. Two values (x/y) indicate \textit{All}/\textit{Overlap} relation experiment settings respectively (if these differ), where \textit{All} indicates experiments with the whole set of biographical relations in each dataset and \textit{Overlap} uses only the intersection. Biographical experiments are performed only with the whole set of biographical relations. $^{\dagger}$Lora Settings: Rank/Alpha/Dropout.}
\label{tab:hyperparams_combined}
\end{table*}

\begin{table*}[ht]
    \centering
    \scriptsize
    \begin{tabular}{lcc|cc|cc}
        \toprule
        & \multicolumn{2}{c}{TACRED-RE} & \multicolumn{2}{c}{NYT} & \multicolumn{2}{c}{Biographical} \\
        \cmidrule(lr){2-3} \cmidrule(lr){4-5} \cmidrule(lr){6-7}
        POS & Head Entity & Tail Entity & Head Entity & Tail Entity & Head Entity & Tail Entity \\
        \midrule
        PROPN  & 77.4 & 55.6 & 98.4 & 98.8 & 87.7 & 60.5 \\
        PRON   & 16.8 & 6.2  & -- & -- & 0.1 & 0.2 \\
        NOUN   & 2.4  & 16.8 & 0.2 & 0.4 & 1.7 & 5.4 \\
        ADJ    & 1.2  & 4.5  & 0.2 & -- & 0.7 & 0.9 \\
        ADP    & 0.7  & 1.6  & 0.2 & 0.3 & 0.3 & 1.1 \\
        NUM    & 0.0  & 8.5  & -- & -- & 6.4 & 25.5 \\
        DET    & 0.3  & 1.0  & 0.2 & 0.1 & 0.8 & 2.3 \\
        VERB   & 0.4  & 0.7  & -- & -- & 0.1 & 0.1 \\
        \bottomrule
    \end{tabular}
    \caption{(Top 8) POS Distribution Across TACRED-RE, NYT, and Biographical Test Sets with all Biographical Relations (\%). POS tags are obtained with spaCy's transformer-based \texttt{en\_core\_web\_trf} model.}
    \label{tab:pos_distribution}
\end{table*}

\begin{table*}[ht]
\section{Misclassification Analysis}
    \centering
    \small
    \renewcommand{\arraystretch}{1.3}
    \begin{tabular}{p{2.2cm}|p{2.5cm}|p{3.8cm}|p{2.3cm}|p{2.2cm}}
        \hline
        \textbf{Issue} & \textbf{Description} & \textbf{Representative Example} & \textbf{Misclassifications on} & \textbf{Models Affected} \\
        \hline
        Overpredicting `None' & Overpredicting `None' and struggling with even clear relations with cues like `born' or`died' & ``My name is \texttt{<e1>}Ruben\texttt{</e1>} and I am from \texttt{<e2>}Holland\texttt{</e2>}'' (GT: \textit{place\_lived}, Pred: \textit{None}; TACRED-RE sample ID: `098f6f318bc468878bbb') & TACRED-RE and Biographical & NYT- and Biographical-adapted models\\
        \hline
        Failure to Capture Implicit Relations & Models struggling to detect implicit relations requiring reasoning & ``\texttt{<e1>}Gross\texttt{</e1>} [...] was sent to \texttt{<e2>}Cuba\texttt{</e2>} as a spy'' (GT: \textit{place\_lived}, Pred: \textit{None}; TACRED-RE sample ID: `098f6f318be29eddb182') & TACRED-RE & NYT-and Biographical- adapted models \\
        \hline
        Expected world knowledge & For NYT and Biographical this issue is also frequently paired with detatable ground truth labels & ``\texttt{<e1>}Augustus\texttt{</e1>} also amassed an impressive art collection and built lavish baroque palaces in Dresden and \texttt{<e2>}Warsaw\texttt{</e2>}'' (GT: \textit{dplace\_name}, Pred: \textit{None}; Biographical sample ID: `mS2/247724') & NYT, TACRED-RE, Biographical & models adapted on all 3 datasets affected \\
        \hline
        Relation Present in Sentence but Not Between Specified Entities & This issue raises concerns about the framing of the RE task itself & ``Jan Malte, [...] resident of \texttt{<e1>}Bridgehampton\texttt{</e1>}, died [...] in \texttt{<e2>}San Francisco\texttt{</e2>}'' (GT: \textit{None}, Pred: \textit{place\_of\_death}; NYT article ID: `/m/vinci8/data1/riedel/projects\allowbreak/relation\allowbreak/kb/nyt1\allowbreak/docstore/nyt-2005-2006.backup\allowbreak/1777142.xml.pb') & NYT, TACRED-RE, Biographical & models adapted on all 3 datasets affected \\
        \hline
        Debatable ground truth (GT) labels & Caused by distantly or semi-supervised manner in which NYT and Biographical were created & ``\texttt{<e1>}Ida Freund\texttt{</e1>} was born in \texttt{<e2>}Austria\texttt{</e2>}'' (GT: \textit{Other}, Pred: \textit{place\_of\_birth}; Biographical sample ID: `mS10/37387826') & TACRED-RE, Biographical & NYT- and Biographical-adapted models  \\
        \hline
        Single-Label Annotation Limitation & Sentences labeled with a single relation may contain additional relations that remain unlabeled & ``\texttt{<e1>}Gross\texttt{</e1>}, who is himself Jewish [...] was sent to \texttt{<e2>}Cuba\texttt{</e2>}'' (GT: \textit{place\_lived}, Pred: \textit{None}; TACRED-RE sample ID: `098f6f318b69f98c850c') & NYT, TACRED-RE, Biographical & models adapted on all 3 datasets affected \\
        \hline
        Relation missing in annotation schema & Lack of granularity needed to fully capture an individual's biography & ``Wen was detained in August and accused of protecting the gang operations masterminded by his sister-in-law, \texttt{<e1>}Xie Caiping\texttt{</e1>}, 46, known as the ``godmother'' of the \texttt{<e2>}Chinese\texttt{</e2>} underworld (GT: \textit{place\_lived}, Pred: \textit{nationality}; TACRED-RE sample ID: `098f637935e6e6d1d093') & NYT, TACRED-RE, Biographical & --- \\
        \hline
        Failure to Capture Relations in Long, Compound Sentences & Models struggling with long-term relational dependencies & ``Ecoffey told jurors that he and another federal agent met with \texttt{<e1>}Graham\texttt{</e1>} in April 1994 in Yellowknife, the city in northwest \texttt{<e2>}Canada\texttt{</e2>} where Graham lived at the time'' (GT: \textit{place\_lived}, Pred: \textit{None}; TACRED-RE sample ID: `098f6f318b3ea9531448') & TACRED-RE & NYT-adapted models \\
        \hline
        
    \end{tabular}
    \caption{Common Misclassification Patterns Across TACRED-RE, NYT, and Biographical.}
    \label{tab:consolidated_misclassification_analysis}
\end{table*}

\begin{table*}[ht]
\centering
\small
\begin{tabular}{p{3cm}|p{3cm}|p{3cm}|p{3cm}}
\hline
\textbf{Relation} & \textbf{NYT} & \textbf{TACRED-RE} & \textbf{Biographical} \\
\hline
None & year, york, united, mr, like, states, president, company, work, city & year, national, president, group, include, state, percent, million, american, china & release, contract, announce, song, star, award, series, role, sign, championship \\
\hline
children & father, son, higgins, clark, favre, richard, mary, daughter, carol, daley & son, daughter, grandchild, survive, wife, year, child, include, gude, jr & daughter, son, child, li, father, mother, wife, give, marry, actor \\
\hline
religion & islam, muhammad, prophet, religion, convert, leader, school, al, church, close & jewish, al, islam, shiite, christian, group, muslim, sunni, mohammed, tantawi & -- \\
\hline
place\_lived & senator, republican, state, year, representative, gov, democrat, city, john, mr & year, state, die, home, york, city, president, live, iran, old & -- \\
\hline
place\_of\_birth & city, year, orleans, chicago, bear, bill, attorney, general, mr, california & bear, grow, family, child, york, year, native, july, old, son & bear, raise, née, grow, family, youth, york, california, city, mother
 \\
\hline
place\_of\_death & die, year, home, city, london, los, angeles, mr, yesterday, paris & die, home, hospital, cancer, paris, wednesday, sunday, find, early, dead & die, home, paris, age, california, near, october, london, live, move
 \\
\hline
\end{tabular}
\caption{Top 10 tokens per overlapping relation in NYT, TACRED-RE, and Biographical datasets, following lemmatisation and stop word removal using spaCy's transformer-based \texttt{en\_core\_web\_trf} model.}
\label{tab:top_words_nyt_tacrev_extended}
\end{table*}

\end{document}